\documentclass[12pt, draftclsnofoot, onecolumn]{IEEEtran}
\usepackage{graphicx}
\usepackage{amssymb}
\usepackage{mathtools}
\usepackage{dsfont}
\usepackage{cite}
\usepackage{stfloats}
\usepackage{subfigure}
\usepackage{psfrag}
\usepackage[mathscr]{euscript}
\usepackage{acronym}  % make an acronym
\usepackage{amsmath}
\usepackage{algorithmicx,algorithm}
\usepackage[noend]{algpseudocode}
\usepackage{booktabs}
\usepackage{bbm}
\usepackage{float}
\usepackage{balance}

\newtheorem{theorem}{$\mathbf{Theorem}$}
\newtheorem{lemma}{$\mathbf{Lemma}$}
\newtheorem{corollary}{$\mathbf{Corollary}$}

\begin{document}

\title{\paperTitle}
\title{Mobility-Aware Cluster Federated Learning in Hierarchical Wireless Networks}
\author{

	    Chenyuan Feng,
	    Howard~H.~Yang, \IEEEmembership{Member, IEEE},
	    Deshun~Hu, 
        Zhiwei Zhao, \IEEEmembership{Member, IEEE},
        Tony~Q.~S.~Quek, \IEEEmembership{Fellow, IEEE},
        and Geyong Min, \IEEEmembership{Member, IEEE}
	    
\thanks{
C.~Feng and  T.~Q.~S.~Quek are with Information Systems Technology and Design, Singapore University of Technology and Design, 487372, Singapore (email: chenyuan\_feng@mymail.sutd.edu.sg, tonyquek@sutd.edu.sg) .}
\thanks{H. H. Yang is with the Zhejiang University/University of Illinois at Urbana-Champaign Institute, Zhejiang University, Haining 314400, China, (email: haoyang@intl.zju.edu.cn).}
\thanks{D. Hu is with the Department of Communication Engineering, Harbin Institute of Technology, Harbin 150001, China (email:hudeshun1993@gmail.com).}
\thanks{Z. Zhao is with the School of Computer Science and Engineering, University of Electronic Science and Technology of China, Chengdu 610051, China (e-mail: zzw@uestc.edu.cn).}
\thanks{G. Min is with the Department of Computer Science, College of Engineering Mathematics and Physical Sciences, The University of Exeter, Exeter EX4 4QF, U.K. (e-mail:  g.min@exeter.ac.uk).}
\thanks{The corresponding author of this paper is H. ~H.~Yang.}
}
\maketitle
\acresetall

\vspace{-50pt}
\begin{abstract}
Implementing federated learning (FL) algorithms in wireless networks has garnered a wide range of attention. However, few works have considered the impact of user mobility on the learning performance. To fill this research gap, firstly, we develop a theoretical model to characterize the hierarchical federated learning (HFL) algorithm in wireless networks where the mobile users may roam across multiple edge access points (APs), leading to incompletion of inconsistent FL training.
Secondly, we provide the convergence analysis of HFL with user mobility. Our analysis proves that the learning performance of HFL deteriorates drastically with highly-mobile users. And this decline in the learning performance will be exacerbated with small number of participants and large data distribution divergences among user's local data. To circumvent these issues, we propose a mobility-aware cluster federated learning (MACFL) algorithm by redesigning the access mechanism, local update rule and model aggregation scheme. Finally, we provide experiments to evaluate the learning performance of HFL and our MACFL. The results show that our MACFL can enhance the learning performance, especially for three different cases: the case of users with non-independent and identical distribution (non-IID) data, the case of users with high mobility, and the cases with a small number of users. 
\end{abstract}
\begin{IEEEkeywords}
Hierarchical federated learning, user mobility, data heterogeneity, convergence analysis
\end{IEEEkeywords}

\section{Introduction}\label{sec:intro}

%background
To support emerging intelligent services and applications,  federated learning (FL) has been proposed in \cite{ref1} as a promising approach to generate high quality models without collecting the distributed data. It provides great convenience for parallel processing and can reduce the costs of message exchanging significantly \cite{ref06,ref07}. As for the learning efficiency, the convergence analysis of the FL in wireless networks has been studied in \cite{ref08,swang2019}, which can characterize the impact of unreliable communication links. To accelerate the convergence of the FL, sophisticated user scheduling schemes in wireless networks have been designed in \cite{Howard2019}, and a theoretical bound has been analyzed in \cite{ref014}. To fully exploit the processing power of cloud and edge computing servers, a client-edge-cloud hierarchical federated learning (HFL) paradigm has been proposed in \cite{CECHFL2020}, and the resource optimization has been designed in \cite{ref021}.

Albeit the popularity, it has been demonstrated that FL yields suboptimal results if the local clients’ data distributions diverge, which causes significant decline of model accuracy \cite{ref2}. Due to the heterogeneous environment, a shared global model trained by conventional FL might not generalize well for each user. To tackle this problem, a new version named cluster multi-task FL is proposed in \cite{cmtl2009} and \cite{mtfl2017}. Cluster multi-task FL can improve the FL framework by dividing clients into different clusters, and different learning tasks and shared models are learned by different clusters. In \cite{FedCluster2020}, the authors have proved that cluster FL will converge faster than single cell FL. Therefore, cluster FL is more suitable for the large-scale hierarchical wireless networks. 

% personalized FL in single cell 
Aside from dividing users into different clusters, another technique named personalized FL is proposed to cope with the data heterogeneity among users. In \cite{perfedavg}, the authors propose the Per-FedAvg algorithm based on a model-agnostic meta-learning formulation. In \cite{pfedme}, the pFedMe algorithm is proposed by adding regularization terms to local loss functions in the version of Moreau envelopes. The authors of \cite{FedAMP} propose the FedAMP algorithm to facilitate the collaboration between similar clients based on attention mechanism. However, most of the works related to personalized FL are formulated for single-cell networks rather than hierarchical wireless networks.

%FL+mobility
In addition to data heterogeneity, user mobility is another important factor in hierarchical wireless networks. However, most of the previous works only consider a static network topology while the impact of user mobility has been ignored. Indeed, in many practical scenarios, one mobile user may download the latest global model from a certain access point (AP) and keep moving and might leave the AP's coverage area during the local training procedure. As far as we know, the state-of-the-art works that considered user mobility in the design of the FL paradigms are \cite{Gmin2020} and \cite{HTN2020}. 
However, in \cite{Gmin2020} and \cite{HTN2020}, the APs do not select the users with high mobility that might leave their coverage area, which can lead to a decrease in the number of participants, hence results in missing valid training data and under-fitting global model.

Although cluster FL with mobile users seems promising and more practical, several essential challenges still exist, which can be listed as follows: \textit{Firstly}, it is difficult to evaluate the impact of user mobility on the convergence rate and accuracy of the HFL. Besides single-cell networks, the convergence rate of multi-layer FL is analyzed in \cite{coopsgd,HLSGD2021,MLSGD2021}. However, there is little work considering the convergence performance of HFL with mobile users. 
\textit{Secondly}, it is difficult to efficiently assign mobile users with non-IID dataset into different clusters with limited communication and computation resources. It is neither efficient for clustering schemes with additional communication and computation costs, nor practical due to the neglect of physical accessibility \cite{CFL2020, FLHC2020,HFL2020,CACFL2020}. 
\textit{Finally}, the existing aggregation schemes are based on the conventional equal-weighted averaging scheme \cite{ref1}, which become the performance bottleneck due to the distribution divergences of non-IID data and user mobility.

%contribution
Motivated by these critical issues, we analyze the the performance of HFL algorithm with mobile users and propose a mobility-aware cluster federated learning (MACFL) algorithm. Our main contributions are summarized as follows:
\begin{itemize}
    \item We develop a theoretical framework for the analysis of HFL with mobile users, whereas convergence rates are derived by accounting for the impacts of user mobility as well as the heterogeneity from both dataset and network architecture.
    \item Building upon the analytical results, we redesign the access mechanism to allow mobile users to effectively participate in collaboration. We also introduce personalized FL and attentive averaging schemes to boost up the convergence rate and enhance model accuracy. 
    \item We conduct extensive experiments to evaluate the performance of our proposed scheme, and also study the impact of network parameters on the learning performance. The experiment results show that our proposed schemes can significantly improve the convergence and accuracy performance of HFL with mobile users.
\end{itemize}

\section{System Model}\label{sec:system}
In this section, we introduce the architecture of the hierarchical wireless network, the mobility model of users, as well as the implementation of FL in such a system. 
 
\begin{figure} \label{fig:cec_HierNet}
    \centering
    \subfigure[]{
   \includegraphics[width=0.44\textwidth]{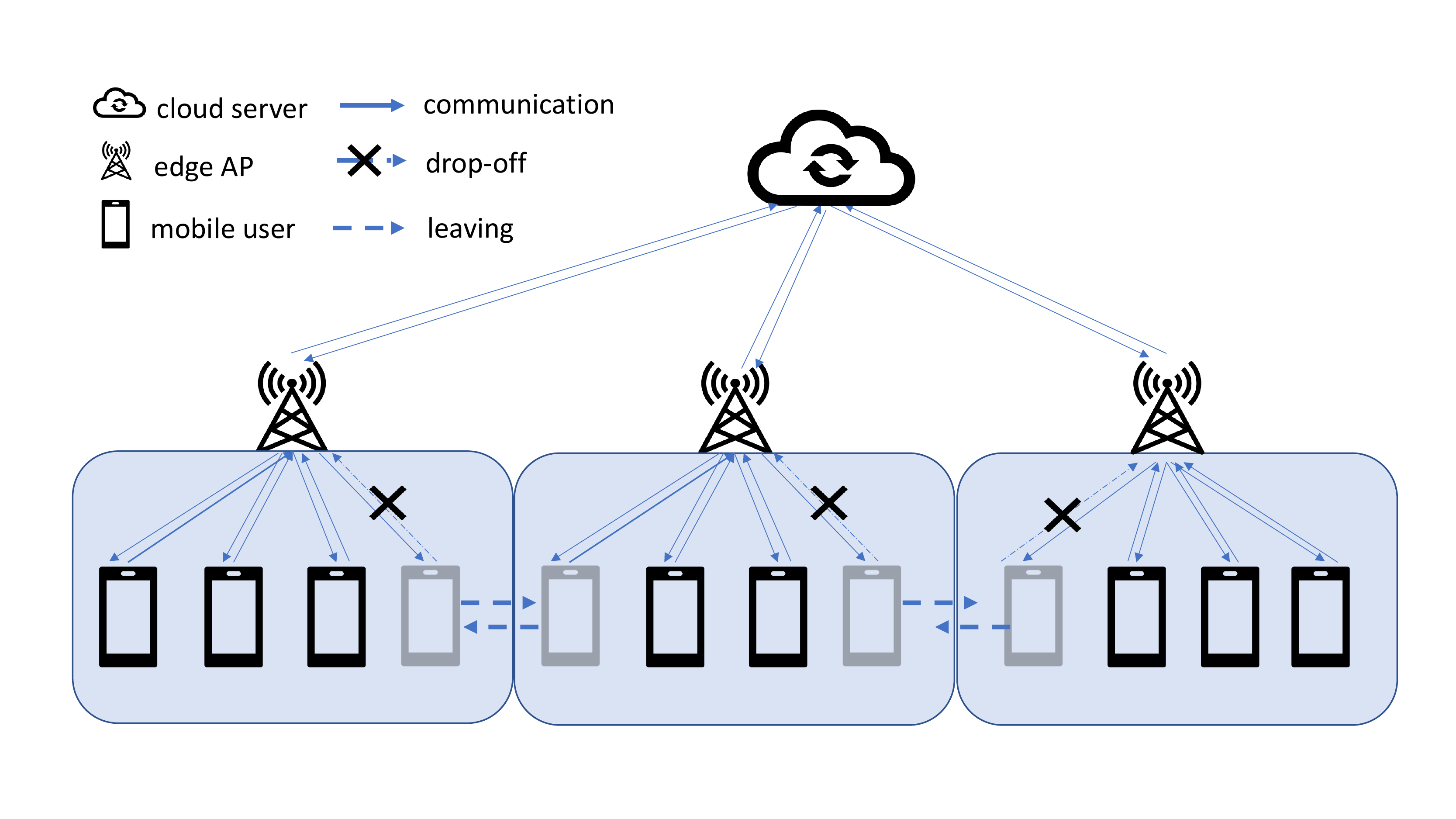}\label{fig:system}}
    \subfigure[]{
   \includegraphics[width=0.5\textwidth]{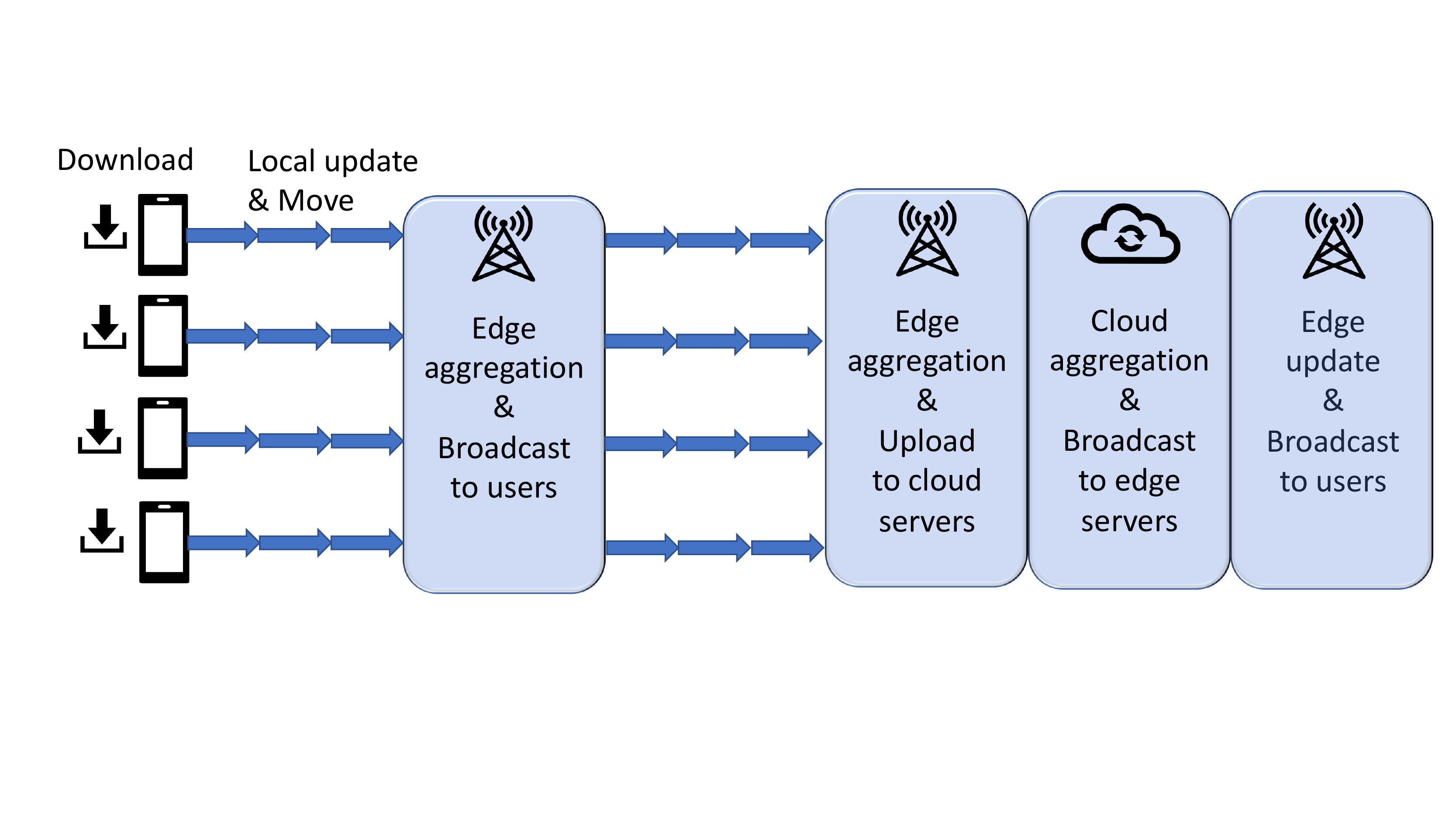}\label{fig:learn}}
    \caption{An illustration of the system model. Fig.\ref{fig:system}: The hierarchical wireless network consists of one cloud server, multiple edge APs, and all users will communicate with their nearest edge AP and might lose connection due to mobility. Fig.\ref{fig:learn} shows an example of the procedure of HFL with $\kappa_1=3$ and $\kappa_2=2$.}
\end{figure}

\subsection{Network Setup}
We consider a statistical learning task conducted in a hierarchical wireless network. 
The network consists of $M$ mobile users, $N$ edge APs, and one cloud server, as depicted in Fig.\ref{fig:system}.  
We denote the set of users by $\{ u_m \}_{m=1}^M$, where a generic user $u_m$ holds a local dataset $\mathcal{D}_{u_m}$. 
The size of dataset $\mathcal{D}_{u_m}$ is denoted as $\vert \mathcal{D}_{u_m} \vert$, where $\vert \cdot \vert$ represents the cardinality of a set. 
We further denote the set of edge APs by $\{ c_n \}_{n=1}^N$, and assume that each edge AP is equipped with an edge server, we interchangeably use the term edge AP and edge server in the rest of paper.
Without loss of generality, we consider $M \gg N$ because an AP is usually connected by multiple users in the real world. 
In this work, we consider a simple cluster scheme based on physical accessibility. Specifically, each mobile user will connect to the nearest edge AP based on real time locations without additional communication and computation overhead. 
As such, we denote $\mathcal{C}_n$ as the set of mobile users connected to the edge AP $c_n$. 

The objective of all the entities in this network is to jointly minimize the global loss functions, $F(\mathbf{w}, \mathcal{D})$, given as follows 
\begin{align}
F(\mathbf{w}, \mathcal{D}) &= \frac{1}{ \vert \mathcal{D} \vert } \sum_{ (\mathbf{x}_i,y_i) \in \mathcal{D} } \ell(\mathbf{w}; \mathbf{x}_i, y_i )
%\nonumber\\&
= \sum_{ m=1 }^M %\frac{ \vert \mathcal{D}_{u_m} \vert }{ \vert \mathcal{D} \vert } 
\alpha_{u_m} F_{u_m}(\mathbf{w},\mathcal{D}_{u_m})
\end{align}
in which $\mathcal{D} = \cup_{m=1}^M \mathcal{D}_{u_m}$ is the dataset aggregated from all the mobile users and $\ell(\mathbf{w}; \mathbf{x}_i, y_i)$ is the loss function assigned on the $i$-th data sample pair $(\mathbf{x}_i, y_i)$. Particularly, $\mathbf{x}_i \in \mathbb{R}^d$ denotes the $i$-th input sample and $y_i$ the corresponding label, and $\mathbf{w}  \in \mathbb{R}^d$ is the minimizer, which can fully parametrize the machine learning model, and $\alpha_{u_m}$ denotes the linear combination weight of user $u_m$ and satisfies $\sum_{m=1}^M \alpha_{u_m} =1$ with $\alpha_{u_m} \in [0,1]$.
Moreover, the function $F_{u_m}(\mathbf{w},\mathcal{D}_{u_m})$ represents the empirical loss function comprised by the local dataset of the mobile user $u_m$, given as
\begin{equation}\label{eq:eq1}
 F_{u_m}(\mathbf{w},\mathcal{D}_m) = \frac{1}{ \vert \mathcal{D}_{u_m} \vert } \sum_{ (\mathbf{x}_i,y_i) \in \mathcal{D}_{u_m} } \ell(\mathbf{w}; \mathbf{x}_i, y_i ).
\end{equation}

Due to privacy concerns of the users, the local dataset is not accessible by neither the APs nor the cloud server. Therefore, the global loss function $F(\mathbf{w},\mathcal{D})$ cannot be directly evaluated and its minimization needs to be carried out by means of federated computing. 
Fig.~\ref{fig:learn} illustrates a typical procedure of HFL. 
Particularly, inside the coverage -- also referred to as cluster -- of each edge AP, model training is conducted between the AP and the mobile users connected to it. 
As such, the objective function of this cluster is given by 
\begin{align}
f_{c_i}(\mathbf{w}_{c_i}, \mathcal{D}_{c_i} ) = \sum_{u_m \in \mathcal{C}_{i} }  \alpha_{u_m}^c F_{u_m}(\mathbf{w},\mathcal{D}_{u_m}) 
\end{align}
where $\mathcal{D}_{c_i} = \bigcup_{u_m \in \mathcal{C}_i} \mathcal{D}_{u_m}$ denotes the the training dataset owned by all users that connects to edge AP $c_i$ and $\alpha_{u_m}^{c} \in [0,1] $ denotes the weight of $u_m$ at edge aggregation in its assigned cluster, whereas $\sum_{m=1}^M \alpha_{u_m}^c=1$. 
Similarly, the edge APs will further aggregate their parameters at the cloud server to minimize the following function
\begin{align}
f(\mathbf{w},\mathcal{D}) = \sum_{i=1}^N \alpha_{c_i} f_{c_i}(\mathbf{w}, \mathcal{D}_{c_i})%= \sum_{m=1}^M \alpha_{u_m} F_{u_m}(\mathbf{w},\mathcal{D}_{u_m})
\end{align}
where $\alpha_{c_i} \in [0,1] $ is the weight of $c_i$ and we have $\sum_{i=1}^N \alpha_{c_i} =1$.  Recalling cluster and global functions, we have $\alpha_{u_m}= \alpha_{u_m}^{c} \alpha_{c_{u_m}}$. 

In this work, we consider the communications between users and edge APs happen every $\kappa_1$ times of local updates. And the communication between the cloud and edge servers happens every $\kappa_2$ times of edge aggregations.  

\subsection{User Mobility Model}
In the presence of user mobility, the set of mobile users associated with each edge AP varies over time. In order to characterize this feature, we assume all the users are uniformly distributed over the entire network at the beginning of time, and then each user will stay or move to a neighboring cluster according to a certain probability during the local training. 

We use a Markov chain to model the dynamics of mobile users. Specifically, we use a vector $\mathbf{\pi}_{u_m}^t\in \mathbb{R}^{N}$ to capture the state of connection for $u_m$ at the $t$-th iteration, where the entries are defined as follows
\begin{equation}
    \pi^{t}_{u_m}[i]=\left\{\begin{matrix}
 1,& \text{if}~u_m \in \mathcal{C}_i^{t}\\ 
 0,& \text{otherwise}\end{matrix}\right.
\end{equation}
in which $\mathcal{C}_i^{t}$ denotes the set of mobile users connected to $c_n$ at the $t$-th iteration. Moreover, we use an adjacency matrix $\mathbf{A} \in \mathbb{R}^{N\times N}$ to characterize the spatial topology between the edge APs. The elements of this matrix are defined %as follows
%\begin{equation}
$    \mathbf{A}_{ij}=\left\{\begin{matrix}
 1,& ~\text{if} ~~ c_j \in \mathcal{N}(c_i)\\ 
 0,& \!\!\!\!\!\!\!\! \text{otherwise}\end{matrix}\right.$
%\end{equation}
where $\mathcal{N}(c_n)$ denotes the set of neighboring APs of $c_n$, including itself, and the size of $\mathcal{N}(c_n)$ is given by $|\mathcal{N}(c_i)|=\sum_{j=1}^N \mathbf{A}_{ij}$.

\begin{figure} 
    \centering
  \includegraphics[width=0.7\textwidth]{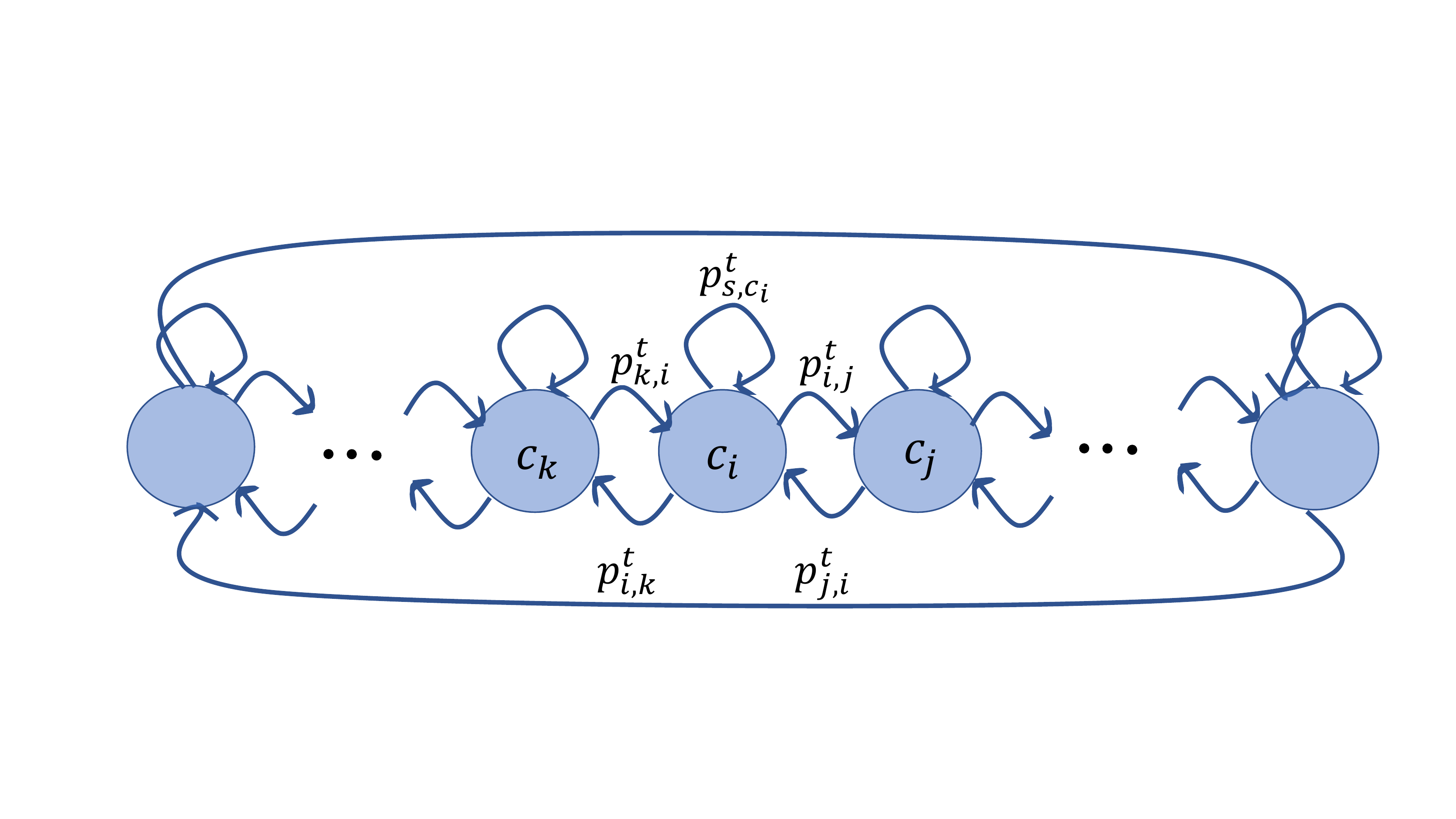}
  \vspace{-10pt}
    \caption{An example of a linear graph and the transmission probability of mobile users between different clusters. } 
    \label{fig:markovnet}
\end{figure}

We assume all the user are uniformly and randomly distributed over the entire network at the beginning of time, and the probability of any specific user moving to different neighboring APs are equal. Therefore, the element of transition probability matrix can be written as 
\begin{equation}
    \mathbf{P}^{t}_{i,j}=\left\{\begin{matrix}
    p_{s,c_i}^t, & \text{if}~ i=j \\ 
    \frac{1-p_{s,c_i}^t}{\left |  \mathcal{N} (c_n) \right |-1},& \text{if} ~~ c_j \in \mathcal{N}(c_i) \setminus c_i \\ 
 0, & \text{otherswise}
\end{matrix}\right.
\end{equation}
If $p_{s,c_i}^t=1, \forall i, \forall t $, the scenario boils down to the conventional HFL with static users. 

In this paper, we assume all the mobile users have the same transition probability matrix. 
Then, at the $t$-th local computation step, let $\mathbf{P}^{t}  \in \mathbb{R}^{N \times N}$ denote the transition probability matrix, which can be used to characterize the trajectory of mobile user at $t$-th iteration. 
As shown in Fig. \ref{fig:markovnet}, the entry at ($i$, $j$), i.e., $\mathbf{P}^{t}_{ij}$, denotes the probability that users staying in AP $c_i$ are moving to $c_j$ at $t$-th local computation time. In our work, we assume the transition probability matrix are the same for all users and each user will stay at the same cluster at $t$-th local computation at the probability of $p_{s,c_i}^t$, and leave for different neighbors at the probability of $1-p_{s,c_i}^t$. Given current observation vector and transmission probability matrix, the future state vector can be estimated by
\begin{equation}
 \pi^{t+1}_{u_m}=\pi^{t}_{u_m}  \mathbf{P}^{t}=\pi^{0}_{u_m}  \prod_{\tau=0}^t  \mathbf{P}^{\tau} 
\end{equation}
Following the above, we can calculate the size $S_i^t$ of the set $\mathcal{C}_i^t$ as follows
\begin{equation}
 S_i^t=\sum_{m=1}^M \pi^{t}_{u_m}[i] = \sum_{m=1}^M \pi^{0}_{u_m}  \left(\prod_{\tau=0}^t P^{\tau}_{u_m}\right)_{[:, i]}
\end{equation}
where $(\cdot)_{[:, i]} $ represents the $i$-th column of a matrix. 
The evolution of $S_i^t$ is a complicated quasi-birth and death process, which depends on the initial conditions, Markov process intensity matrix, and the number of iterations \cite{MBDP2018}. For the sake of tractability, we consider an equilibrium condition with steady-state probability distribution as follows: 
\begin{equation}\label{eq:sizeassumption1}
 S_i^t (1-p_{s,c_i}^t) = \sum_{c_j\in \mathcal{N}(c_i) \setminus c_i } S_j^t p_{c_j,c_i}^t, \forall i, \forall t.  
\end{equation}
This equation indicates the number of incoming and departing users in each cluster is balanced, then we have $S_i^t =S_i, \forall t, \forall i$. It is noteworthy that the indexes of user in $\mathcal{C}_i^t$ are time varying albeit the size keeps the same. 

\section{HFL with Mobile Users}\label{sec:hfl}
In this section, we extend the conventional HFL algorithm \cite{CECHFL2020} to account for cases with mobile users.
The general scheme is given in Algorithm \ref{alg:1}, whereas the key steps, as well as the convergence analysis, are elaborated in the sequel. 

\subsection{Algorithm Description}
\begin{algorithm}[t!]
	\caption{Conventional HFL with Mobile Users}
	\label{alg:1}
	\begin{algorithmic}[1]
	\State \textbf{Initialize} $\{\mathcal{D}_{u_m}\}_{m=1}^M$,  $\{\pi_{u_m}^0\}_{m=1}^M$, $\mathbf{P}$, $\mathcal{C}^0=\{\mathcal{C}_1^0,...,\mathcal{C}_N^0\}$, ${\{\mathbf{w}_{c_i}^{(0)}\}}_{i=1}^N$. 
\For{each edge communication round $b=0,1,...,B-1$}
		 \For{each user $m=1,2,...,M$ in parallel}
		  \State{$t=b\kappa_1$}
		     \State{Local update based on \eqref{eq:local_e1} and \eqref{eq:localdl1}.} 
	           \EndFor
		   \For{each edge $n=1,....,N$ in parallel}
		      \State{Edge update based on  \eqref{eq:edgedl2}.}		
		   \EndFor
		   \If{$i \mod \kappa_2 = 0$}
		       \State{Cloud update based on \eqref{eq:clouddl3}.}
		       \For{each edge $n=1,....,N$ in parallel}
		             \State{$\mathbf{w}_{c_n}^{t+1}  \leftarrow \mathbf{w}_g^{t+1}$}
		       \EndFor
		   \EndIf
\EndFor
\State \Return Global model $\mathbf{w}_g^T$
\end{algorithmic}
\end{algorithm}

The key steps of the HFL algorithm are presented in Algorithm \ref{alg:1}.  During the local training procedure, each mobile user will download the cluster model from its closest AP and then update local model after the local training. Only mobile users who stay in the same coverage area at the local training procedure can upload successfully to the original edge AP, and the cloud server will make an aggregation of all cluster models from all the edge APs’ models after every $\kappa_2$ edge model aggregations. The procedures of user update, edge update, and cloud update are detailed in below, respectively. 

\subsubsection{Local Update}
Let $t = b \kappa_1 + j$ denote the index of local update iteration, with $i$ denoting the index of edge communication round and $j$ denoting the index of epoch during the local training. $\mathbf{w}^{t}_{u_m}$ denotes the local model parameter of $u_m$ at the $t$-th local update iteration, when $t \mod \kappa_1 =0 $, each user downloads the latest cluster model from its tagged edge AP, i.e.,
\begin{equation}\label{eq:local_e1}
    \mathbf{w}^{t}_{u_m} =\mathbf{w}^{t}_{c_n} , \text{ if } \pi_{u_m}^{t}[n]=1,
\end{equation}
and then performs $\kappa_1 \geq 1$ steps of local updates via SGD methods, which can be expressed as
\begin{align}
\label{eq:localdl1}
   &\mathbf{w}^{t+1}_{u_m} \leftarrow \mathbf{w}^{t}_{u_m}-\eta g(\mathbf{w}^{t}_{u_m},\xi^{t+1}_{u_m}), 
\end{align}
where $\eta$ is the learning rate, $\xi^{t}_{u_m} \subset \mathcal{D}_{u_m}$ is a mini-batch independently and identically sampled from the local dataset at the $t$-th local update iteration, and $g(\mathbf{w}^{t}_{u_m},\xi^{t}_{u_m}) $ denotes the mini batch gradient of local loss function, and it satisfies $\mathbb{E}_{\xi} \{g(\mathbf{w}^{t}_{u_m},\xi^{t}_{u_m})\} = \nabla F(\mathbf{w}^{t}_{u_m},\mathcal{D}_{u_m})$. 
For notation simplicity, we use $g(\mathbf{w}^{t}_{u_m})$ and $\nabla F (\mathbf{w}^{t}_{u_m})$ for short in the rest of this paper. 

\subsubsection{Edge update}
The communications between the users and their associated edge APs take place in every $\kappa_1$ local update iterations. 
Due to mobility, the set of mobile users in cluster $c_n$ at $b$-th edge communication round might be different from that at the $(b+1)$-th edge communication round. 
As such, the edge AP only collects the updated gradients from the users that are connect to $c_n$ at the $(b+1)$-th edge aggregation can upload successfully. 
Let a Boolean variable $\mathbb{I}_{u_m}^{b\kappa_1}$ represent the uploading state, i.e., $\mathbb{I}_{u_m}^{b\kappa_1}=1$ if  $\pi_{u_m}^{(b+1)\kappa_1}[i]=\pi_{u_m}^{b\kappa_1} [i],\forall i $, otherwise $\mathbb{I}_{u_m}^{b\kappa_1}=0$. Then the cluster model is updated as follows:
\begin{equation}\label{eq:edgedl2}
 \mathbf{w}_{c_i}^{(b+1)\kappa_1} =  \mathbf{w}_{c_i}^{b\kappa_1}- \eta \sum_{u_m \in \mathcal{C}_i^{t}} \sum_{\tau=1}^{\kappa_1} \alpha_{u_m}^c g(\mathbf{w}^{b\kappa_1+\tau}_{u_m}) \mathbb{I}_{u_m}^{b\kappa_1}
  \end{equation}
where $ \mathbf{w}_{c_i}^{b\kappa_1}$ and $ \mathbf{w}_{c_i}^{(b+1)\kappa_1}$ are the cluster models of the edge AP $c_i$ after the $b$-th and $(b+1)$-th edge communication round, respectively. 
If $b \mod  \kappa_2 \neq 0$, the edge AP will broadcast the latest cluster model to its connected users. And after every $\kappa_2$ edge aggregations, the edge APs will upload the cluster models to the cloud server,  and then update the cluster model after receiving the global model from the cloud, and finally broadcast to the connected users. 

\subsubsection{Cloud Update}
The global aggregation happens in every $\kappa_2$ edge aggregations, which means the global model is updated as follows when $b \mod \kappa_2 = 0$:
\begin{equation}\label{eq:clouddl3}
\mathbf{w}_g^{b\kappa_1} =  \sum_{n=1}^N \alpha_{c_n}  \mathbf{w}_{c_n}^{b\kappa_1}
\end{equation}
where $\alpha_{c_n}$ denotes the weight of updated cluster models from $c_n$. Since the global aggregation happens every $\kappa_1 \kappa_2$ local update iterations, we also have 
\begin{equation}\label{eq:clouddl31}
\mathbf{w}_g^{(b+\kappa_2)\kappa_1} = \mathbf{w}_g^{b\kappa_1}- \eta \sum_{m=1}^M \sum_{q=0}^{\kappa_2-1} \sum_{\tau=1}^{\kappa_1} \alpha_{u_m} g(\mathbf{\tilde{w}}^{(b+q)\kappa_1+\tau}_{u_m}) \mathbb{I}_{u_m}^{(b+q)\kappa_1}
\end{equation}
Compared to the time spent in local training, the time duration of parameter uploading, model aggregation, and downloading is relatively short. Therefore, we assume the connection states of the mobile users will only change during the local training procedures and remain the same during communication and model aggregation procedures. 

\subsection{Convergence Analysis}
Owing to mobility, users may drop off from the associated edge APs, which leads to a decrease in the number of participants in the FL. 
In this section, we analyze the impact of user mobility on the convergence rate of the HFL.  
Similar to previous works, we make the following assumptions:
\begin{itemize}
\item { \textit{The loss functions are lower-bounded: } $f(\mathbf{w}) \geq f_{\text{inf}}, \forall \mathbf{w}$.}
\item {\textit{The gradients of loss functions are bounded: } $\| \nabla F(\mathbf{w})\|^2 \leq G^2, \forall \mathbf{w}$.}
\item {\textit{The loss functions are $L$-smooth:} $\| \nabla F(\mathbf{w}_1)-\nabla F(\mathbf{w}_2)\| \leq L \|\mathbf{w}_1-\mathbf{w}_2\|, \forall \mathbf{w}_1, \mathbf{w}_2$.}
\item {\textit{The mini-batch gradients are unbiased with bounded variance: } $\mathbb{E}_{\xi |\mathbf{w}} \{g(\mathbf{w})\} = \nabla  F(\mathbf{w})$, ${ \| g(\mathbf{w}) - \nabla F(\mathbf{w}) \| }^2 \leq  \sigma^2, \forall \mathbf{w}, \xi$. }
\item {\textit{The divergences of the local, cluster and global loss functions are bounded, for all $u_m, c_i, \mathbf{w}$, we have  } $\frac{1}{N} \sum_{i=1}^{N}  { \|  \nabla f_{c_i}(\mathbf{w}) - \nabla f(\mathbf{w}) \| }^2  \leq  \epsilon_g^2$, $\frac{1} {S_i }\sum_{u_m \in \mathcal{C}_i } { \|  \nabla F_{u_m}(\mathbf{w}) - \nabla f_{c_i}(\mathbf{w}) \| }^2  \leq  \epsilon_c^2.$}
\end{itemize}

Although the averaged global model is not observable when $t \mod \kappa_1 \kappa_2 \neq 0$ in this system, here we use $\mathbf{\bar{w}}_g^t$ for analysis similar to related work \cite{HLSGD2021, MLSGD2021}. 
Because the objective function $F (\mathbf{w},\mathcal{D})$ can be non-convex, the learning algorithm via SGD methods may converge to a local minimum or saddle point. 
Similar to \cite{HLSGD2021,MLSGD2021} and \cite{Lian2015,Bottou2018}, we leverage the expected gradient norm as an indicator of convergence, namely the training algorithm achieves an $\theta$-suboptimal solution if $\mathbb{E}\left[ \frac{1}{T}\sum_{t=1}^T {\| \nabla f(\mathbf{\bar{w}}_g^t)\|}^2 \right] \leq \theta$, where $\theta > 0$. This condition can guarantee the convergence of an algorithm to a stationary point.

\begin{lemma}\label{lemma1}
\textit{
For the employed FL system, when the learning rate satisfies $\eta  <  \frac{1}{L}$, after $T$ rounds of global aggregations, we have
\begin{align}\label{eq:lemma1}
\theta_{HFL} \leq & \frac{2}{\eta p_s T}  \big[\, \mathbb{E} f (\mathbf{\bar{w}}^0_g) -f_{\inf}) \, \big] + \eta L  \sigma^2 \sum_{m=1}^M (\alpha_{u_m})^2   +  4\epsilon_g^2 + 4 \epsilon_c^2\nonumber\\ 
 & + \frac{4 L^2}{T}\sum_{t=0}^{T-1}\bigg( \sum_{i=1}^N \alpha_{c_n} \mathbb{E} \Big[ \Vert \mathbf{\bar{w}}^{t}_g - \mathbf{\bar{w}}_{c_i}^t \Vert^2 \Big] +   \sum_{m=1}^M \alpha_{u_m} \mathbb{E} \Big[ \Vert \mathbf{\bar{w}}_{c_{u_m}}^t - \mathbf{w}_{u_m}^t \Vert^2 \Big] \bigg).
\end{align}
} 
\end{lemma}
\begin{IEEEproof}
Please refer to Appendix \ref{ap:lemma1}.
\end{IEEEproof}
Notably, the first term on the right hand side of \eqref{eq:lemma1} is similar to the setting of centralized learning \cite{Bottou2018} when $p_s=1$, and the second term is introduced by the randomness arises from the mini-batch gradients, the third and fourth terms correspond to the divergences among the local, cluster, and global loss functions, the fifth and sixth terms are introduced by the divergences among local, cluster, and global model parameters. 
When $p_s=0$, the right hand of the inequality of \eqref{eq:lemma1} is unbounded, which means the training algorithm does not converge if no mobile user is participating in the model aggregation.

We further bound the mean square error (MSE) of the local, as well as cluster, model parameters in below.
\begin{lemma}\label{lemma2}
\textit{
For the employed FL system, if $\eta < \frac{1}{\sqrt{12}L\kappa_1}$, the MSE of the local model parameters can be bounded as follows: 
 \begin{align}\label{eq:lemma2}
 & \frac{1}{T} \sum_{t=0}^{T-1} \sum_{m=1}^M  \alpha_{u_m} \mathbb{E} \| \mathbf{w}^{t}_{u_m}- \mathbf{\bar{w}}_{c_{u_m}}^t \|^2 \nonumber\\
 & \leq   \frac{ 1 } {1-12 \eta^2 L^2 \kappa_1^2 p_s^2}  \bigg( 6 \eta^2  p_s^2 \kappa_1^2 \epsilon_c^2 + 2 \eta^2 \kappa_1  p_s \left(\sigma^2+(1-p_s) G^2\right) \sum_{m=1}^M \alpha_{u_m} (1-\alpha_{u_m}^c)    \bigg). 
 \end{align}
 }
 \end{lemma}
 
\begin{IEEEproof}
Please refer to Appendix~\ref{ap:lemma2}.
\end{IEEEproof}

\begin{lemma}\label{lemma3}
\textit{
For the employed FL system, if $\eta < \frac{1}{\sqrt{12}L\kappa_1\kappa_2}$, the MSE of the cluster model parameters can be bounded as follows: 
\begin{align}\label{eq:lemma3}
&  \frac{1}{T} \sum_{t=0}^{T-1}\sum_{i=1}^N \alpha_{c_n}  \mathbb{E} \| \mathbf{\bar{w}}^{t}_g - \mathbf{\bar{w}}_{c_i}^t \|^2 \nonumber\\
&\leq   \frac{  1 } {1-12\eta^2L^2 \kappa_1^2 \kappa_2^2 p_s^2 } \bigg\{ 6 \eta^2 p_s^2 \kappa_1^2 \kappa_2^2  \epsilon_g^2 
+ 4 \eta^2 \kappa_1\kappa_2 p_s ( \sigma^2 + (1-p_s) G^2) \sum_{m=1}^M \alpha_{u_m}(\alpha_{u_m}^c-\alpha_{u_m}) \nonumber\\
& + \frac{ 4 \eta^2 p_s^2 L^2  \kappa_1^2 \kappa_2^2  } {1-12 \eta^2 L^2 \kappa_1^2 p_s^2 } \bigg[ 2 \eta^2 \kappa_1  p_s (\sigma^2+(1-p_s) G^2) \sum_{m=1}^M \alpha_{u_m} (1-\alpha_{u_m}^c)  + 6 \eta^2  p_s^2 \kappa_1^2 \epsilon_c^2   \bigg] \bigg\}
 \end{align}
 }
\end{lemma}

\begin{IEEEproof}
Please refer to Appendix \ref{ap:lemma3}.
\end{IEEEproof}

We are now in position to obtain the upper bound of $\theta_{HFL}$ by applying Lemma \ref{lemma2} and Lemma \ref{lemma3} back into Lemma \ref{lemma1}. 
\begin{theorem}\label{theorem1}
\textit{
Under the employed FL system, when the learning rate is chosen as $\eta < \frac{1}{\sqrt{12}L\kappa_1\kappa_2}$, then after $T$ rounds of global aggregations, Algorithm~1 achieves an $\theta_{HFL}$-suboptimal solution, where $\theta_{HFL}$ is bounded as  
\begin{align}\label{eq:theorem1}
\theta_{HFL}
& \leq \frac{2}{\eta p_s T}  \big[ \mathbb{E} f (\mathbf{\bar{w}}^0_g) -f_{\inf} \big] +\eta L \sigma^2 \sum_{m=1}^M \alpha_{u_m}^2  +  \frac{ 4(1- 6 L^2 \eta^2 p_s^2 \kappa_1^2 \kappa_2^2 ) } {1-12\eta^2L^2 \kappa_1^2 \kappa_2^2 p_s^2 } \epsilon_g^2  \nonumber\\
& + 4 \bigg[1+ \frac{  ( 1-8 \eta^2 p_s^2 L^2  \kappa_1^2 \kappa_2^2 ) 6 \eta^2  L^2 p_s^2 \kappa_1^2 } { (1-12\eta^2L^2 \kappa_1^2 \kappa_2^2 p_s^2 )(1-12 \eta^2 L^2 \kappa_1^2 p_s^2)} \bigg] \epsilon_c^2   + \frac{  8 L^2 \eta^2 \kappa_1  p_s (\sigma^2+(1-p_s) G^2) } {1-12\eta^2L^2 \kappa_1^2 \kappa_2^2 p_s^2 }   \nonumber\\ 
& \times  \sum_{m=1}^M \alpha_{u_m} \bigg[ 2 \kappa_2 ( \alpha_{u_m}^c -  \alpha_{u_m}) + \frac{ 1-8 \eta^2 p_s^2 L^2  \kappa_1^2 \kappa_2^2 } { 1-12 \eta^2 L^2 \kappa_1^2 p_s^2}  (1- \alpha_{u_m}^c) \bigg].
 \end{align}
 }
\end{theorem}
From the right hand side of \eqref{eq:theorem1}, we can see that the convergence error floor depends on the initial states, divergences between local, cluster, and global loss functions, SGD variance, and rate of descent of loss functions. 

\textbf{Remark 1:}{
\textit{By taking the partial derivatives with respect to $\kappa_1$ and $\kappa_2$, respectively, we find that $\theta_{HFL}$ monotonously goes down with a decreasing value of $\kappa_1$ or $\kappa_2$, which implies the more often the users and edge APs exchange information, the faster the training algorithm converges. 
Additionally, if the product of $ \kappa_1 \kappa_2$ is fixed, $\theta_{HFL}$ will decline with a decrease in $\kappa_1$ or an increase in $\kappa_2$. 
This observation suggests exchanging local models frequently helps more than exchanging global models, when the duration of global aggregation is fixed.
}
}

\textbf{Remark 2:}{
\textit{By taking a partial derivative with respect to $p_s$, we can see that $\theta_{HFL}$ is not a monotonic function of  $p_s$ which depends on the properties of training data and loss functions employed by local training. 
Particularly, when $p_s=0$, the algorithm will never converge since there is no mobile user participating in the model aggregation, which means a new model aggregation scheme needs to be devised for users with high mobility. }
}

In addition, by recalling the definition of $\epsilon_g^2$ and $\epsilon_c^2$, it is clear that the divergences of  local and cluster loss functions with IID data is smaller than those with non-IID data. 

\begin{corollary}
\textit{
Under the employed FL system, if the weights for parameter aggregations are chosen as $\alpha_{c_n}=  \frac{1}{N}$, $\alpha_{u_m} = \frac{1}{M}$, and  $\alpha_{u_m}^c =  \frac{N}{M}$, and the learning rate is set as $\eta < \frac{1}{\sqrt{12}L\kappa_1\kappa_2}$, then after $T$ rounds of global aggregations, Algorithm~\ref{theorem1} can achieve an $\theta_{HFL}$-suboptimal solution, where $\theta_{HFL}$ is bounded as  
\begin{align}\label{eq:coro1}
\theta_{HFL}
& \leq \frac{2 (\mathbb{E} f (\mathbf{\bar{w}}^0_g) -f_{\inf})}{\eta p_s T}  + \frac{\eta L}{M} \sigma^2 \nonumber\\
& + \frac{ 4(1- 6 L^2 \eta^2 p_s^2 \kappa_1^2 \kappa_2^2 ) } {1-12\eta^2L^2 \kappa_1^2 \kappa_2^2 p_s^2 } \epsilon_g^2  + 4 \bigg[1+ \frac{  ( 1-8 \eta^2 p_s^2 L^2  \kappa_1^2 \kappa_2^2 ) 6 \eta^2  L^2 p_s^2 \kappa_1^2 } { (1-12\eta^2L^2 \kappa_1^2 \kappa_2^2 p_s^2 )(1-12 \eta^2 L^2 \kappa_1^2 p_s^2)} \bigg] \epsilon_c^2      \nonumber\\ 
&  + \frac{  8 L^2 \eta^2 \kappa_1  p_s (\sigma^2+(1-p_s) G^2) } {1-12\eta^2L^2 \kappa_1^2 \kappa_2^2 p_s^2 } \bigg[  2 \kappa_2 \frac{N-1}{M} + \frac{ 1-8 \eta^2 p_s^2 L^2  \kappa_1^2 \kappa_2^2 } { 1-12 \eta^2 L^2 \kappa_1^2 p_s^2}   \frac{M-N}{M}\bigg].
 \end{align}
 }
\end{corollary}

Following Corollary~1, we can see that if $M=N$, there is $\epsilon_c^2=0$ and $\alpha_{u_m}^c =  1$, if $N=1$, we have $\epsilon_g^2=0$, $\kappa_2=1$ and  $\alpha_{u_m}^c =  \alpha_{u_m} $, the case turns into single-cell FL.  If $M=N=1$, we have $\theta_{HFL}
\leq \frac{2}{\eta p_s T}  (\mathbb{E} f (\mathbf{\bar{w}}^0_g) -f_{\inf}) +\eta L \sigma^2$, which is in accord with centralized learning with SGD method \cite{Bottou2018} with $p_s=1$. 

%%%%%%%%%%%%%%%%%%%%%%%%%%%%%%%%%%%%%%%%%%%%%%%%
% our proposed model
%%%%%%%%%%%%%%%%%%%%%%%%%%%%%%%%%%%%%%%%%%%%%%%%
\section{Mobility-Aware Cluster federated learning}
\label{sec:macfl}

In this section, we propose the MACFL algorithm by redesigning the access mechanism, personalized local model update rule, and weighted average schemes. Specifically, our proposed MACFL algorithm allows a mobile user to download a cluster model from one edge AP and, if it roams to another cell, upload the parameter to the corresponding AP, which leads to more mobile users to successfully participate in model aggregations. Based on the analysis in Theorem ~\ref{theorem1}, the shared cluster models learned by conventional HFL can not generalize well due to the user mobility and data heterogeneity. Therefore, we also introduce personalized FL and attentive weighted averaging schemes to reduce the performance degradation caused by SGD variance and the divergences among local, cluster, and global loss functions. In this sections, we apply these state-of-the-art schemes to HFL with mobile users and analyze its convergence rate.

\subsection{Learning Task}

In our proposed MACFL algorithm, the newly-arrived users can update their local models based on the cluster model downloaded by one edge AP, and then upload the updated results to a different edge AP. This mechanism leads to an increasing number of participants. However, considering user mobility, the shared cluster model aims to learn a common model for the time-varing user set, and does to adapt to each user. Inspired by the personalized FL in single-cell networks such as Per-FedAvg \cite{perfedavg}, we propose a novel personalized cluster FL for hierarchical wireless networks to learn shared cluster and global models for each server and personalized local models for each mobile user at the same time. Moreover, motivated by attention-based FL in single-cell networks such as FedAMP \cite{FedAMP}, we introduce an attentive collaboration into HFL to boost the collaboration effectiveness between users without leakage of private data, which can also mitigate the impact of user mobility and data heterogeneity.

In our work, we aim to learn personalized local and cluster models and shared gloabl model at the same time, similar to \cite{perfedavg}, we first rewrite the global loss function in the following way
\begin{align}\label{eq:p2}
&\tilde{f}( \{\mathbf{w}_{u_m}\}_{m=1}^M,\mathbf{w}_{g}, \mathcal{D}) = \sum_{m=1}^M \beta_{u_m} \tilde{F}_{u_m}(\mathbf{w}_{u_m},\mathbf{w}_{g},\mathcal{D}_{u_m}), \nonumber\\
& \tilde{F}_{u_m}(\mathbf{w}_{u_m},\mathbf{w}_{g},\mathcal{D}_{u_m}) = F_{u_m}( (\mathbf{w}_g-\rho \nabla F_{u_m}( \mathbf{w}_{u_m},\mathcal{D}_{u_m})), \mathcal{D}_{u_m})
\end{align}
where $\tilde{f}( \{\mathbf{w}_{u_m}\}_{m=1}^M,\mathbf{w}_{g}, \mathcal{D})$ and $\tilde{F}_{u_m}(\mathbf{w}_{u_m},\mathbf{w}_{g},\mathcal{D}_{u_m})$ denote the global and local loss function in MACFL algorithm with $F_{u_m}$  defined as \eqref{eq:eq1}, and $\beta_{u_m}$ is the weight assigned on the local models learned by  $u_m$ at global aggregation. And the global model $\mathbf{w}_{g}$ is obtained by collaborative learning of all users, and the personalized local model is learned by each user based on the shared model and one more step of gradient decent, which can be computed as $\mathbf{w}_g-\rho \nabla F_{u_m}( \mathbf{w}_{u_m},\mathcal{D}_{u_m})$ with $\rho$ denoting the stepsize of local training.

\subsection{Algorithm Description}
\subsubsection{Local Update}
With a loss function given as \eqref{eq:p2}, at the $b$-th edge communication round, each user will at first download the latest personalized model from nearest edge AP, i.e.,  $\mathbf{w}^{b \kappa_1}_{u_m}  = \mathbf{w}^{b \kappa_1}_{c_n}$ if $u_m \in \mathcal{C}_n^{b\kappa_1}$, and then perform $\kappa_1 \geq 1$ steps of local updates via SGD. 
Let $t = b\kappa_1+j, j=0,..., \kappa_1-1$, the evolution of each iteration can be expressed as 
\begin{align}\label{eq:cfl_localup}
&\mathbf{w}^{t+1}_{u_m}= \mathbf{w}^{t}_{u_m} -\eta \tilde{g}(\mathbf{w}^{t}_{u_m}), %\nonumber\\& 
~ \text{with }\tilde{g}(\mathbf{w}^{t}_{u_m}) = (\mathbf{I}-\rho \nabla g(\mathbf{w}^{t}_{u_m})) g(\mathbf{w}^{t}_{u_m}-\rho g (\mathbf{w}^{t}_{u_m}) )
\end{align}
where $\tilde{g}(\mathbf{w}^{t}_{u_m})$ denotes the mini-batch gradient of  local loss function $\tilde{F}_{u_m}(\mathbf{w}^{t}_{u_m})$ given in \eqref{eq:p2} at $i$-th local update after $t$-th edge comunication round, $\mathbf{I}$ denotes the identity matrix, and $\nabla g(\mathbf{w}^{t}_{u_m})$ denotes the gradient of mini-batch gradient $g(\mathbf{w}^{t}_{u_m})$ given in \eqref{eq:localdl1}. As states in \cite{perfedavg}, the gradient estimates can be approximated by first-order approximations with a slight loss of accuracy in order to avoid the high computation complexity of computing hessian term, namely, $\tilde{g}(\mathbf{w}^{t}_{u_m}) \approx g(\mathbf{w}^{t}_{u_m}-\rho g (\mathbf{w}^{t}_{u_m}) )$. 

\subsubsection{Edge Update}
In conventional FL, the aggregation strategy usually weights the importance of each local model equally or proportional to the size of the local data. However, this averaging scheme might not be the true weight of local data of each user in mixture data distribution of cluster and global model.  Moreover, the mobile user sets in each cluster changes every communication round, the new users who might have different cluster model at last iteration, therefore a new edge update method needs to be designed. As shown in \cite{ref2}, the performance of model aggregation depends on a reasonable design of weight coefficients. %Each user, AP and cloud can be viewed as a graph node, the model obtained by each node can be viewed as a graph signal, and the procedure of model aggregation can turn into a signal passing problem in graph learning field, which can learn weight coefficients and new graph signals at the same time via graph attention network \cite{GAT2018}. Compared with other clustered FL with deterministic weight, the weight coefficients are learnable parameters in our work.
 Motivated by the attention scheme employed in FedAMP\cite{FedAMP}, we redesign the edge update rule for $(b+1)$-th edge communication round as follows
\begin{equation}\label{eq:cfl_edgeup}
   \mathbf{w}_{c_n}^{(b+1)\kappa_1} = \sum_{u_m \in \mathcal{C}_n^{(b+1)\kappa_1}} \beta_{u_m}^c  \mathbf{w}^{(b+1)\kappa_1}_{u_m},
~ \text{with }\beta_{u_m}^c = \frac{e^{-\sigma_1 cos(\mathbf{w}_{u_m}^{(b+1)\kappa_1},\mathbf{w}_{c_n}^{b\kappa_1})}}{\sum_{u_m \in \mathcal{C}_n^{(b+1)\kappa_1} } e^{-\sigma_1 cos(\mathbf{w}_{u_m}^{(b+1)\kappa_1},\mathbf{w}_{c_n}^{b\kappa_1})}}, 
 \end{equation}
where $\beta_{u_m}^c$ is the attention coefficient, $\sigma_1$ is a scalar hyper-parameter, the attention coefficient $\beta_{u_m}^c$ denotes the linear combination weight of model parameter aggregating from user $u_m$ to AP $c_n$,  and $\cos(\mathbf{w}_{u_m}^{(b+1)\kappa_1},\mathbf{w}_{c_n}^{b\kappa_1}) $ is the cosine similarity between $\mathbf{w}_{u_m}^{(b+1)\kappa_1}$ and $\mathbf{w}_{c_n}^{b\kappa_1}$ which is defined as $\cos(\mathbf{x},\mathbf{y}) = \frac{<\mathbf{x},\mathbf{y}>}{\|\mathbf{x} \|^2 \| \mathbf{y}\|^2} $. In this work, we employ a simple but effective attention mechanism with only one scalar hyper-parameter. Compared with other clustered FL with deterministic weight, the weight coefficients are learnable parameters in our work. There are other complex methods to learn attention coefficients, such as neural networks \cite{GAT2018}, which will result in a larger computation overhead. Future work will focus on sophisticated attention scheme with limited computation resources.
After edge update, the edge server will distribute the cluster model to connecting users, and when it is time for cloud update, the cluster model will be sent to the cloud server. 

\subsubsection{Cloud Update}
When $b \mod \kappa_2 =0$, the cloud server will collect the cluster models from all the edge APs and update the global parameter as follows
\begin{align}\label{eq:cfl_cloudup}
    {\mathbf{w}}_{g}^{b\kappa_1} = \sum_{n=1}^N  \beta_{c_n} {\mathbf{w}}^{b\kappa_1}_{c_n},
~\text{with }\beta_{c_n} = \frac{e^{-\sigma_2 cos(\mathbf{w}_{c_n}^{b\kappa_1},\mathbf{w}_{g}^{(b-\kappa_2)\kappa_1})}}{\sum_{n=1 }^N  e^{-\sigma_2 cos(\mathbf{w}_{c_n}^{b\kappa_1},\mathbf{w}_{g}^{(b-\kappa_2)\kappa_1})}}
\end{align}
where  $\beta_{c_n}$ is the attention coefficient,  $\sigma_2$ is a scalar hyper-parameter. 

After cloud update, the cloud server will send latest global model to all edge APs, and all edge servers will distribute the latest shared model to its connecting users for a new round of local computing. 

\subsection{Convergence Analysis}
By invoking similar approaches in Section~III, we can analyze the convergence rate of the proposed training scheme.
\begin{theorem}\label{2theorem}
\textit{
For the employed FL system, if the learning rate is chosen as $\eta \leq \frac{1}{\sqrt{12}L \kappa_1 \kappa_2}$, then after $T$ rounds of global aggregations, Algorithm~2 can achieve an $\epsilon_{MACFL}$-suboptimal solution, where $\epsilon_{MACFL}$ is bounded as follows
\begin{align}
\theta_{MACFL}  
& \leq \frac{2}{\eta T}  \big[ \mathbb{E} \tilde{f} (\mathbf{\bar{w}}^0_g) - \tilde{f}_{\inf} \big] + \eta L  \sigma_M^2 \sum_{m=1}^M \beta_{u_m}^2  \nonumber\\ 
& +  \frac{  12 L^2 \eta^2  \kappa_1^2 \kappa_2^2} {1-12\eta^2L^2 \kappa_1^2 \kappa_2^2 }  \epsilon_{M,g}^2 
+\frac{ 12 L^2   \eta^2  \kappa_1^2(1-8 \eta^2  L^2  \kappa_1^2  \kappa_2^2 )  } {(1-12 \eta^2 L^2 \kappa_1^2)(1-12 \eta^2 L^2 \kappa_1^2 \kappa_2^2) }   \epsilon_{M,c}^2  \nonumber\\
& +   \frac{4 L^2   \eta^2  \kappa_1  } {1-12 \eta^2 L^2 \kappa_1^2 \kappa_2^2 }   \sigma_M^2 \sum_{m=1}^M \beta_{u_m} 
\bigg( \frac{1-8 \eta^2  L^2  \kappa_1^2  \kappa_2^2} {1-12 \eta^2 L^2 \kappa_1^2 }(1-\beta_{u_m}^c) + 2 \kappa_2 (\beta_{u_m}^c-\beta_{u_m})\bigg).
 \end{align}
 }
\end{theorem}
\begin{IEEEproof}
Please refer to Appendix \ref{ap:2theorem}.
\end{IEEEproof}

\section{Performance Evaluation}
\label{sec:simulation}
In this section, we conduct experimental evaluation of the cluster FL algorithm on heterogeneous models and dataset configurations to verify the efficacy of our analysis and proposed scheme. 

\subsection{Experimental Settings}
In our experiments, we consider a FL system consisting of 50 users and 5 clusters formed by the APs and one cloud server, and the 5 clusters constitute a linear graph as shown in Fig. \ref{fig:markovnet}.
All users are uniformly and randomly assigned to 5 clusters at the initial time.
We consider an image classification learning task based on the standard MNIST \cite{mnist} dataset, which consists of 60,000 images for training and 10,000 images for testing and contains 10 different hand-written digits. 
At the user side, we consider the Convolutional Neural Network (CNN) models with cross-entropy loss functions for local training. 
The mini-batch size at each SGD step is set as 10, and the learning rate is set as 0.001. 
We set the total iterations to be 2000. 

Because data heterogeneity is critical in FL,  we employ two configurations of data distributions: IID and non-IID dataset cases. 
The size of local dataset owned by each user is set as $|\mathcal{D}_{u_m}|=600$ for both IID and non-IID dataset cases. As for the IID dataset case,  we assume the training dataset of each user is an IID sample of the global training dataset. As for the non-IID dataset case,  we consider the pathological non-IID setting in \cite{ref1}, in which each user only has at most two categories of digits. 

\subsection{Impact of User Mobility on Conventional HFL}
We first carry out experiments to assess the convergence rate of the HFL algorithm with mobile users to illustrate the impact of user mobility with different system parameters. Prior to giving out the results, we would like to note that on the one hand, user mobility leads to a decreasing number of participants at each model aggregation round, which may result in missing training data and under-fitting training model. On the other hand, mobility of the users also has the potential to shuffle the data, which helps reducing the divergences among clusters and construct a representative sample at each SGD step of global model. 

\subsubsection{Impact of the Frequency of Aggregation} 
\begin{figure}
    \centering
  \subfigure[]{
  \includegraphics[width=0.45\textwidth]{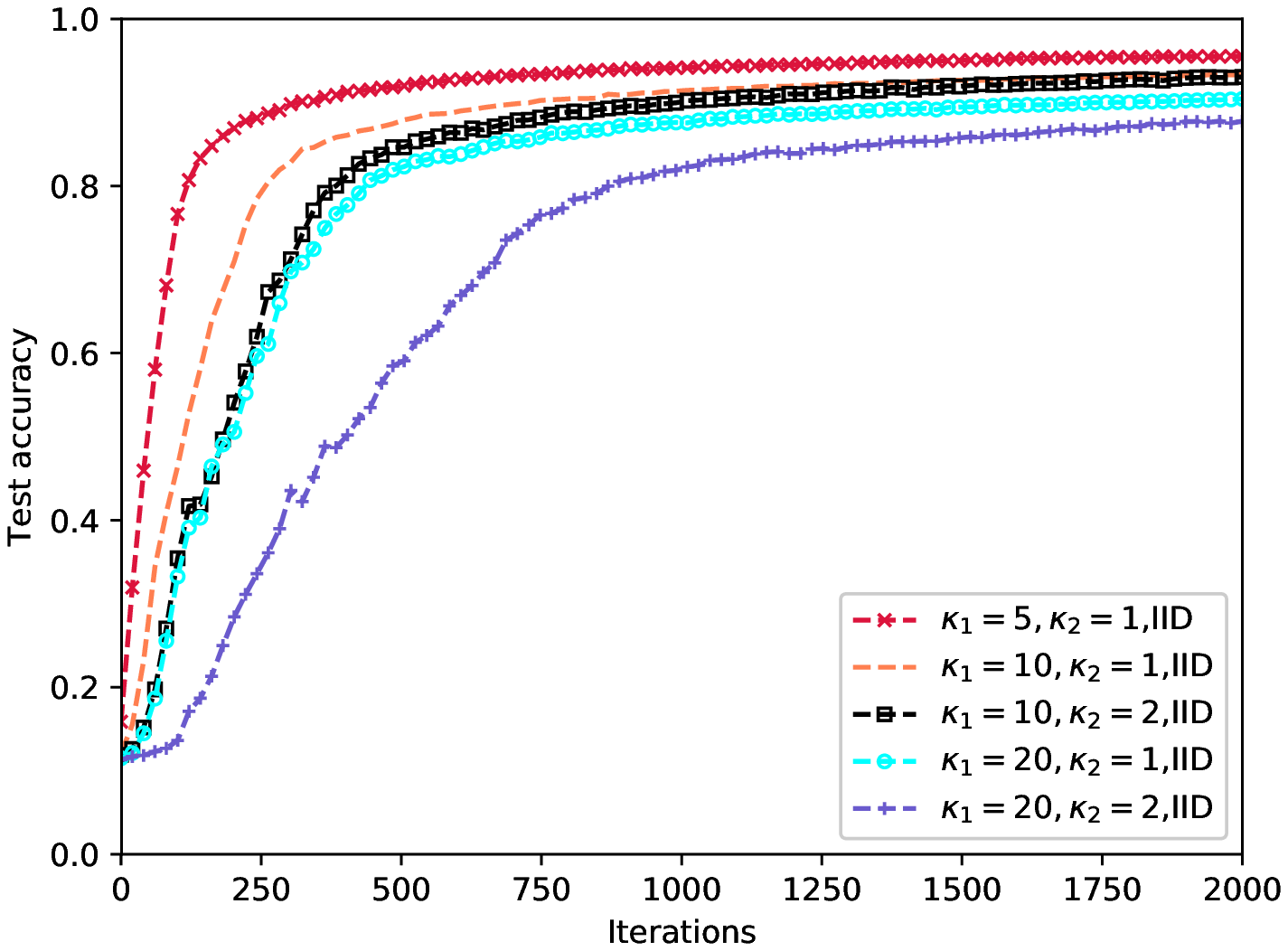} }
    \subfigure[]{
  \includegraphics[width=0.45\textwidth]{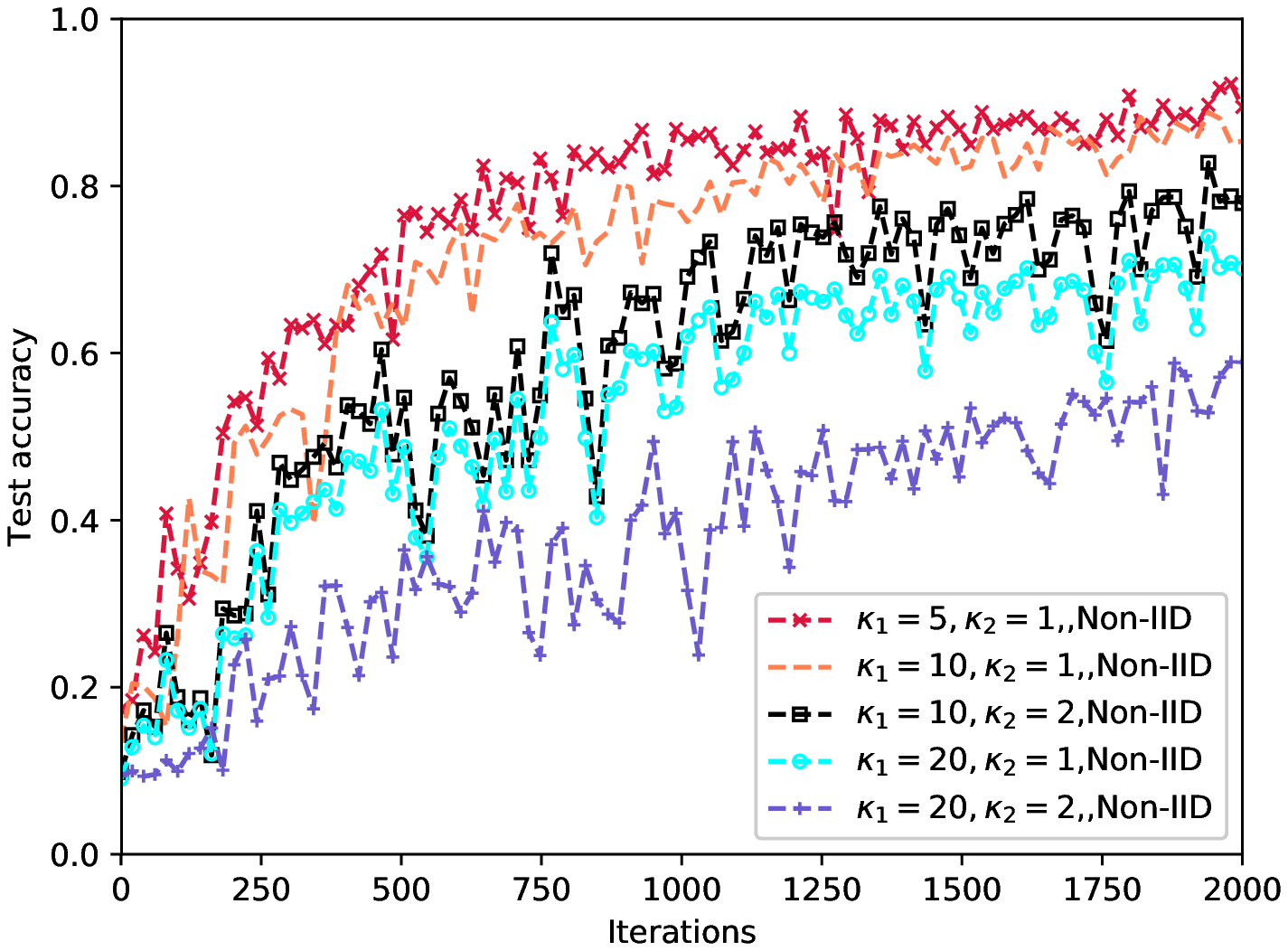} }
    \caption{Evaluation of test accuracy performance with different  $\kappa_1,\kappa_2$, when $p_s=0.5, M=50$ and $N=5$. }
     \label{fig:ta_difk1k2}
\end{figure}

Fig.~\ref{fig:ta_difk1k2} demonstrates the impact of parameter aggregation intervals, $\kappa_1$ and $\kappa_2$, on the accuracy of the trained model. 
It is worthwhile to note that $\kappa_1$ and $\kappa_2$ characterize the frequencies of edge aggregation and global aggregation, respectively, and there are $\kappa_2$ times of communication rounds between users and edge servers and one communication round between cloud and edge servers during one cloud update procedure. 
From Fig.~\ref{fig:ta_difk1k2} we can see that the convergence rates under IID and non-IID dataset can be boosted up by reducing $\kappa_1$ or $\kappa_2$, i.e., increasing the frequency of communications between either the users and edge APs or the edge APs and cloud server. 
Additionally, if we fix the product of $\kappa_1$ and $\kappa_2$, the convergence rate is accelerated with decreasing $\kappa_1$, which suggests that allowing more frequent edge aggregations helps to enhance learning performance. 
These observations verify the convergence analysis given in Theorem \ref{theorem1} and accord with related multi-layer FL settings \cite{HLSGD2021} and \cite{MLSGD2021}. 
This result poses a dissent on the conclusions drawn from single-cell FL \cite{ref1}, where the test accuracy can be improved by increasing $\kappa_1$. 
The main reason attributes to the fact that increasing $\kappa_1$ will enlarge the divergences among local models and increasing $\kappa_2$ will also aggravate the weight difference among cluster models. Similar to \cite{MLSGD2021}, our results show that high frequency of local average, instead of large number of local update iterations,  is beneficial to the learning performance in HFL. 
Moreover, it can be seen that the learning performance of IID dataset case is better than that of non-IID dataset case, which also coincides with the conclusions indicated by Theorem \ref{theorem1}. This is because the value of the divergences of local and cluster loss functions and the variance of SGD of IID dataset case is smaller than those of non-IID dataset case. 

\subsubsection{Impact of Staying Probability} 
\begin{figure}    
    \centering
   \subfigure[]{
  \includegraphics[width=0.45\textwidth]{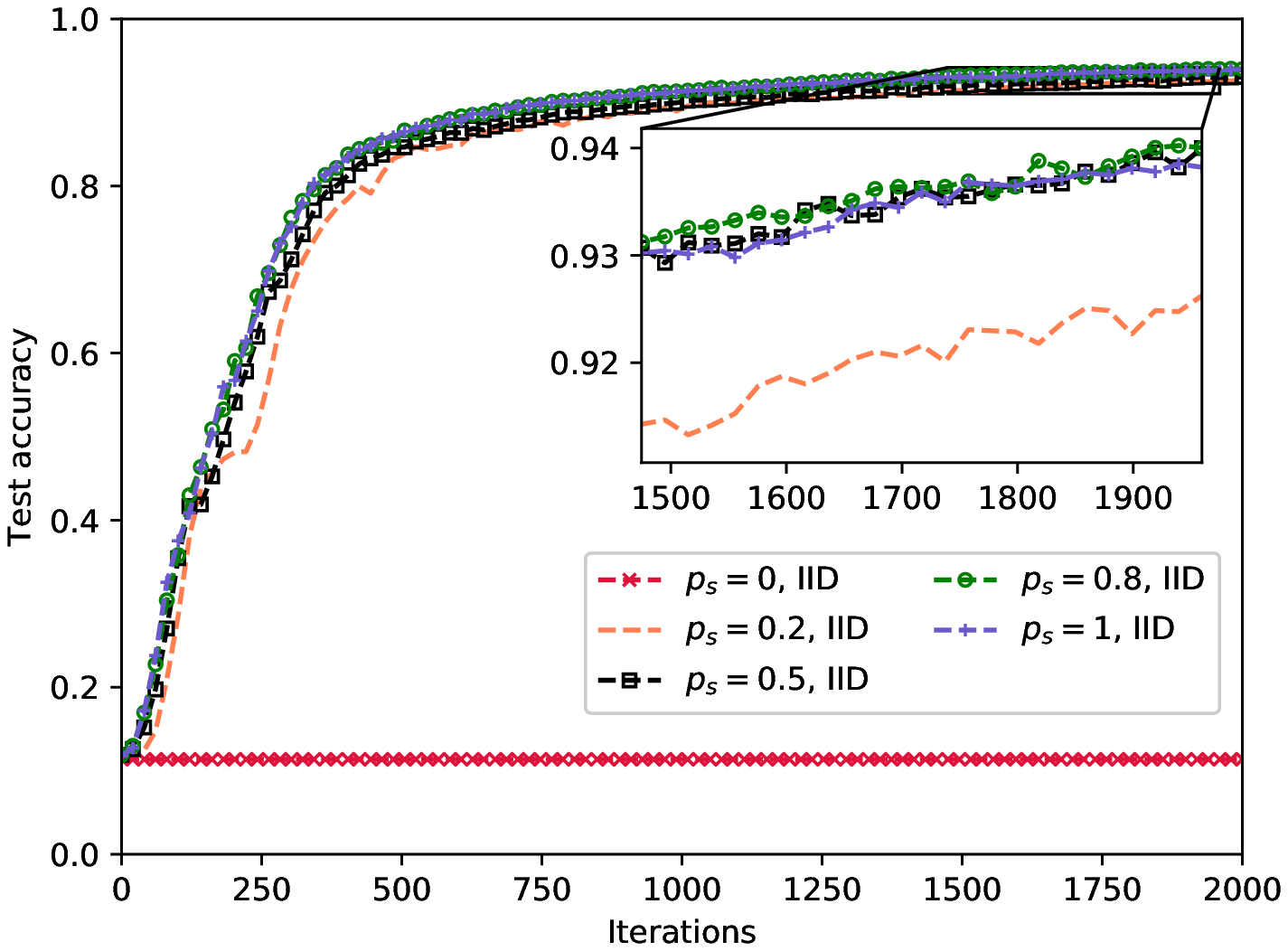} }
    \subfigure[]{
  \includegraphics[width=0.45\textwidth]{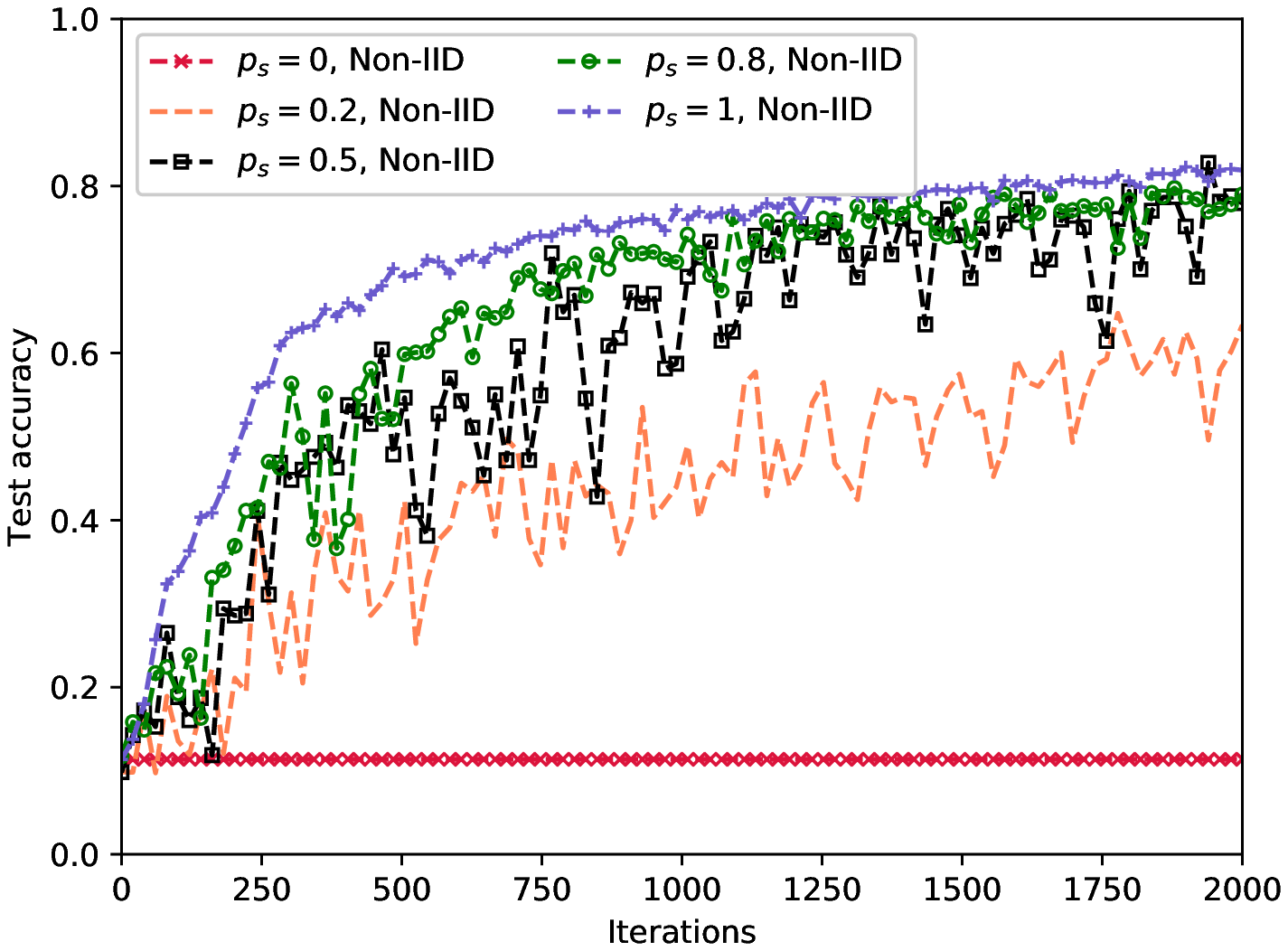} }
    \caption{Evaluation of test accuracy performance with different  $p_s$ when $\kappa_1=20,\kappa_2=1, M=50$ and $N=5$. }
    \label{fig:ta_difps}
\end{figure}

In Fig.~\ref{fig:ta_difps}, we illustrate the impact of user staying probability $p_s$ on the convergence rate of HFL algorithm. 
From this figure, we first notice that the convergence rate changes slightly for IID dataset case with different $p_s > 0$, which implies that the user mobility has a mild effect on the learning performance as long as a portion of the mobile users can participate in the model aggregation. 
This is because the test accuracy is insensitive to the number of participants for the ideal FL with IID dataset in both single-cell \cite{ref1} and multi-cell \cite{coopsgd} networks as long as the mini-batch sample is a representative IID sample of the entire dataset. 
Therefore, the test accuracy will not deteriorate severely by reducing $p_s$ as long as $p_s \neq 0$.
For non-IID dataset case, we observe that increasing $p_s$ can bring along a marked gain in the learning performance. 
This comes from the fact that when dataset is non-IID, it is more likely that a large number of participants contribute a representative sample at each model aggregation step \cite{ref1,coopsgd}. 
Therefore, increasing $p_s$ retains more effective participants in each cluster, which can enhance the learning performance with non-IID datasets.
Moreover, this staying probability also has a relation to the impact of $\kappa_1$ in Fig.~\ref{fig:ta_difk1k2} in real life. Particularly, a small $\kappa_1$ means a shorter duration of local update, and thus the staying probability during local update becomes larger, which helps to speed up HFL for non-IID dataset case. Additionally, increasing $p_s >0$ also helps to reduce the vibrations of the convergence curves, especially for non-IID dataset case. 
Two special scenarios are also noteworthy: When $p_s=0$, there are no updated results collected successfully by edge servers due to high mobility, and the test accuracy merely depends on the initialization model for both IID and non-IID dataset cases. When $p_s=1$, it turns to the conventional HFL with static users. And the test accuracy with $p_s=1$ for IID dataset case is higher comparing with non-IID dataset case, which is reasonable. 

\subsubsection{Impact of the Number of Users} 
\begin{figure}    
    \centering
   \subfigure[]{
  \includegraphics[width=0.45\textwidth]{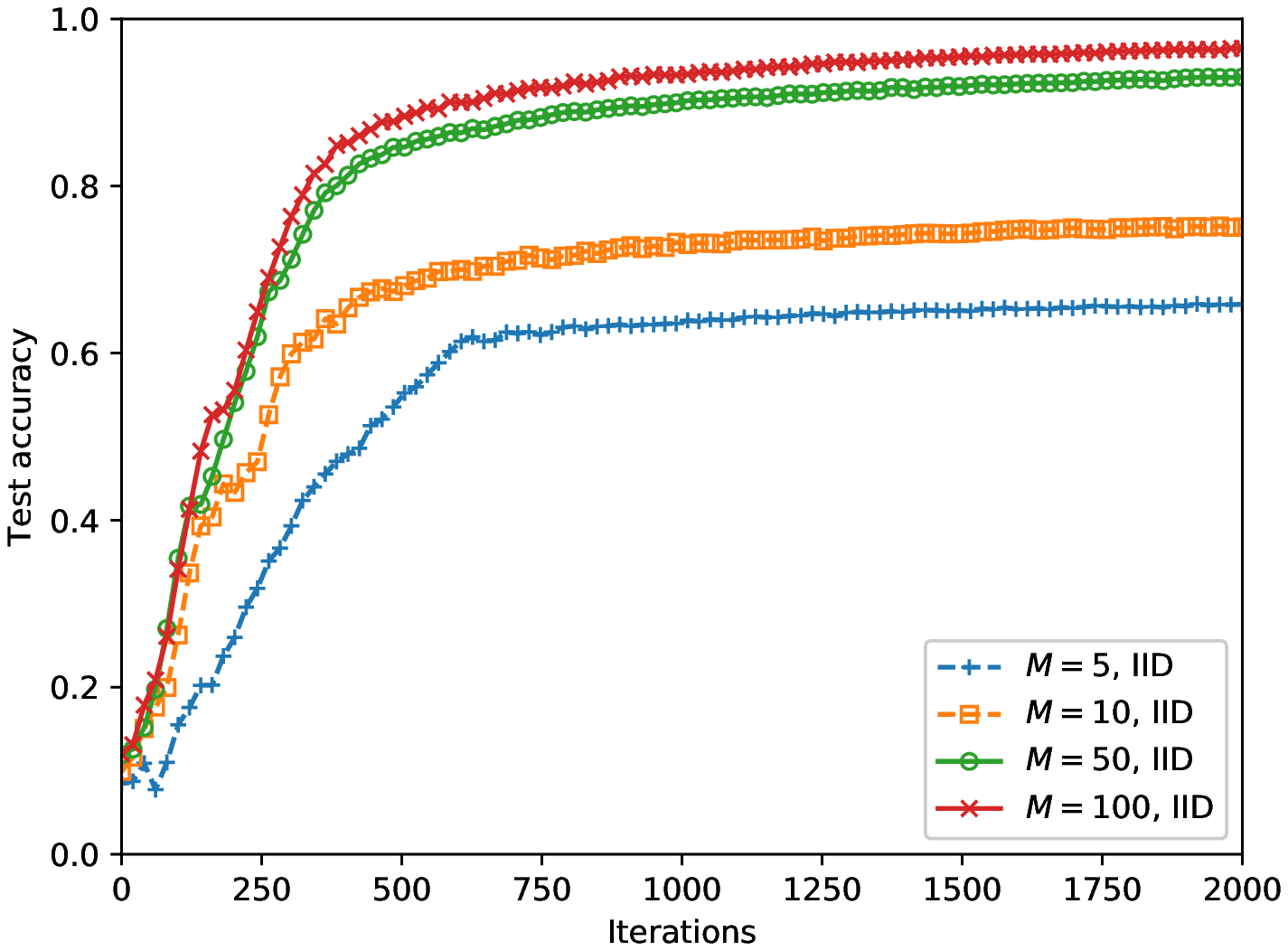} }
    \subfigure[]{
  \includegraphics[width=0.45\textwidth]{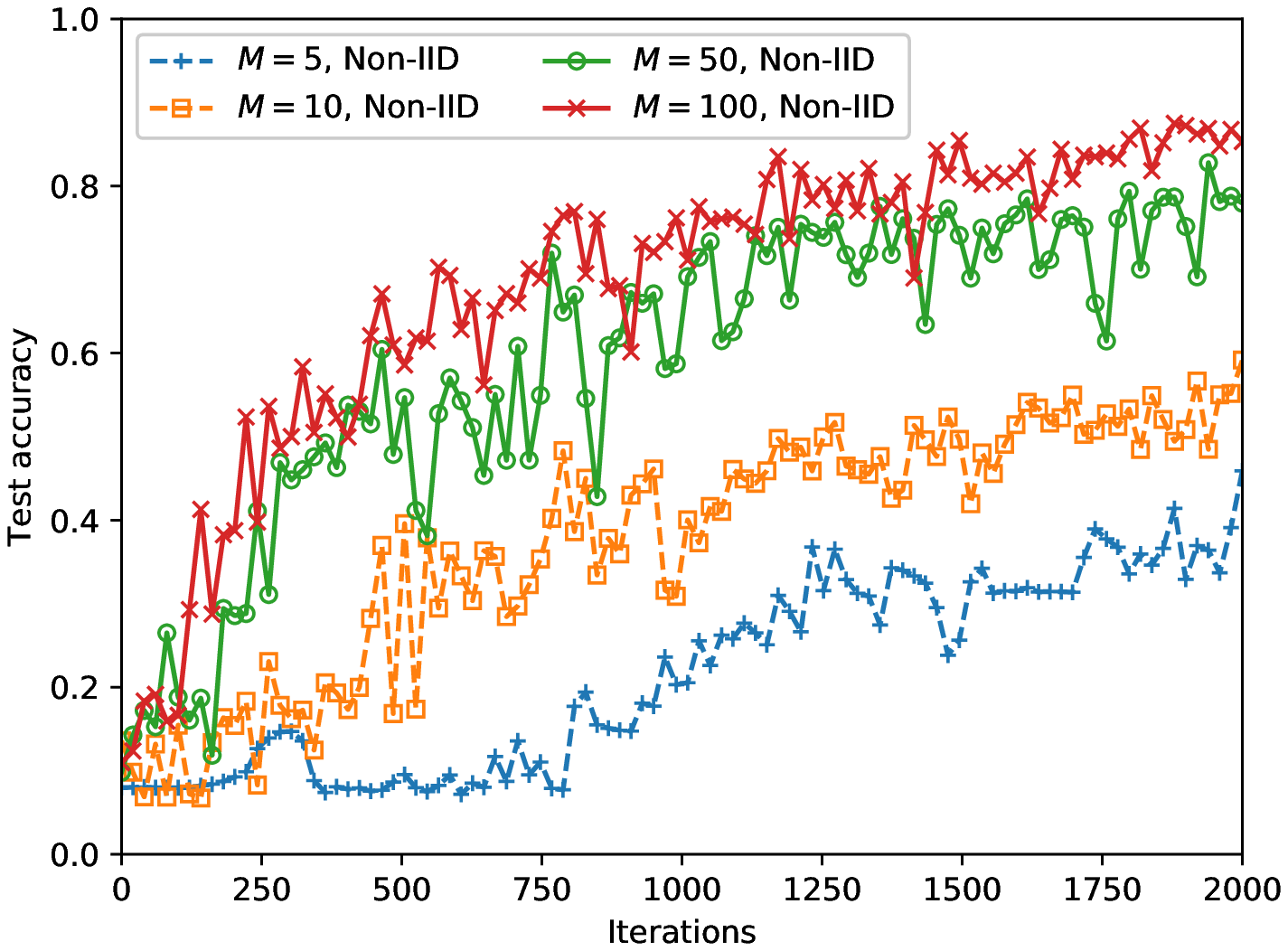} }
    \caption{Evaluation of test accuracy performance with different  $M$ when $p_s=0.5, \kappa_1=20,\kappa_2=1$ and $N=5$. }
    \label{fig:ta_difm}
\end{figure}
Fig.~\ref{fig:ta_difm} depicts the convergence rate of HFL under different user number $M$.  
It can be seen that the convergence rate increases with respect to $M$ for both IID and non-IID dataset cases, which is in accord with convergence analysis in conventional FL \cite{HLSGD2021} and \cite{Bottou2018}. 
Additionally, increasing $M$ also helps to reduce the vibrations of the convergence curves, especially when the number of total global aggregations is small. It suggests that increasing the number of participating users benefits the learning performance when $T$ is small for IID dataset.
This result confirms our analysis given in Theorem ~\ref{theorem1}, namely, updates via SGD from a larger number of users will reduce the variance of SGD at each iteration, which is also in line with \cite{ref6}. 

\subsection{Comparison between HFL and Our Proposed MACFL Algorithm}
We now turn our attention to the performance of the proposed MACFL. 
In this part, we perform several comparisons between our proposed MACFL and the conventional HFL algorithm. 
For a fair comparison, we adopt the same system and common hyperparameters, namely $\eta$, $\kappa_1$, $\kappa_2$, and $M$ for both algorithms. We also use the same initial state for both two algorithms. 
For the hyperparameters only used in MACFL algorithm, namely, $\sigma_1$, $\sigma_2$, and $\rho$, we conduct a grid search to figure out the combination of fine-tuned parameters via learning from the experiences of Per-FedAvg \cite{perfedavg} and FedAMP \cite{FedAMP}. And in our simulation we set $\sigma_1=\sigma_2=25$ and $\rho=0.001$ for both IID and non-IID datasets.

\begin{figure}
    \centering
       \subfigure[]{
  \includegraphics[width=0.45\textwidth]{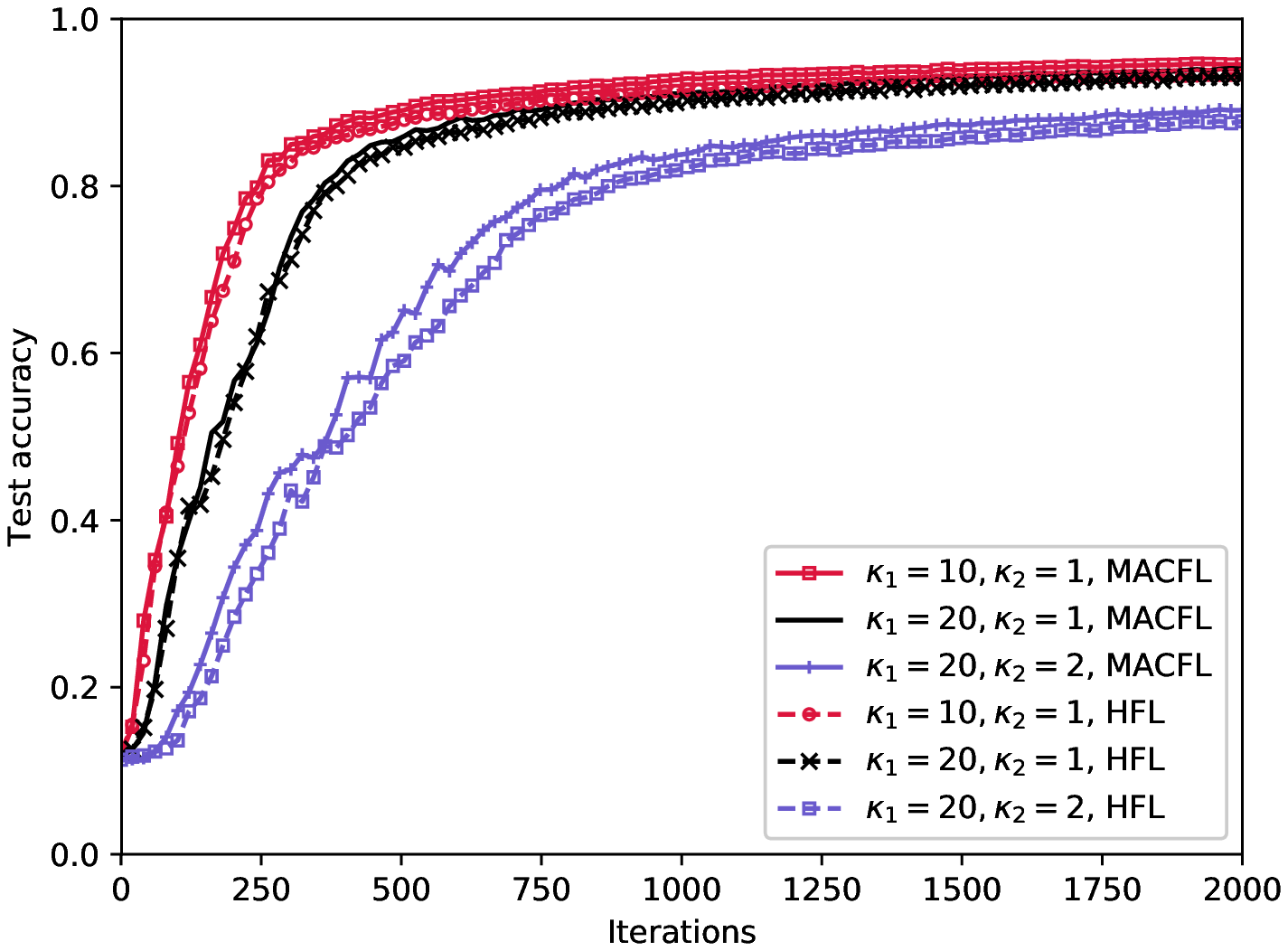}}
    \subfigure[]{
  \includegraphics[width=0.45\textwidth]{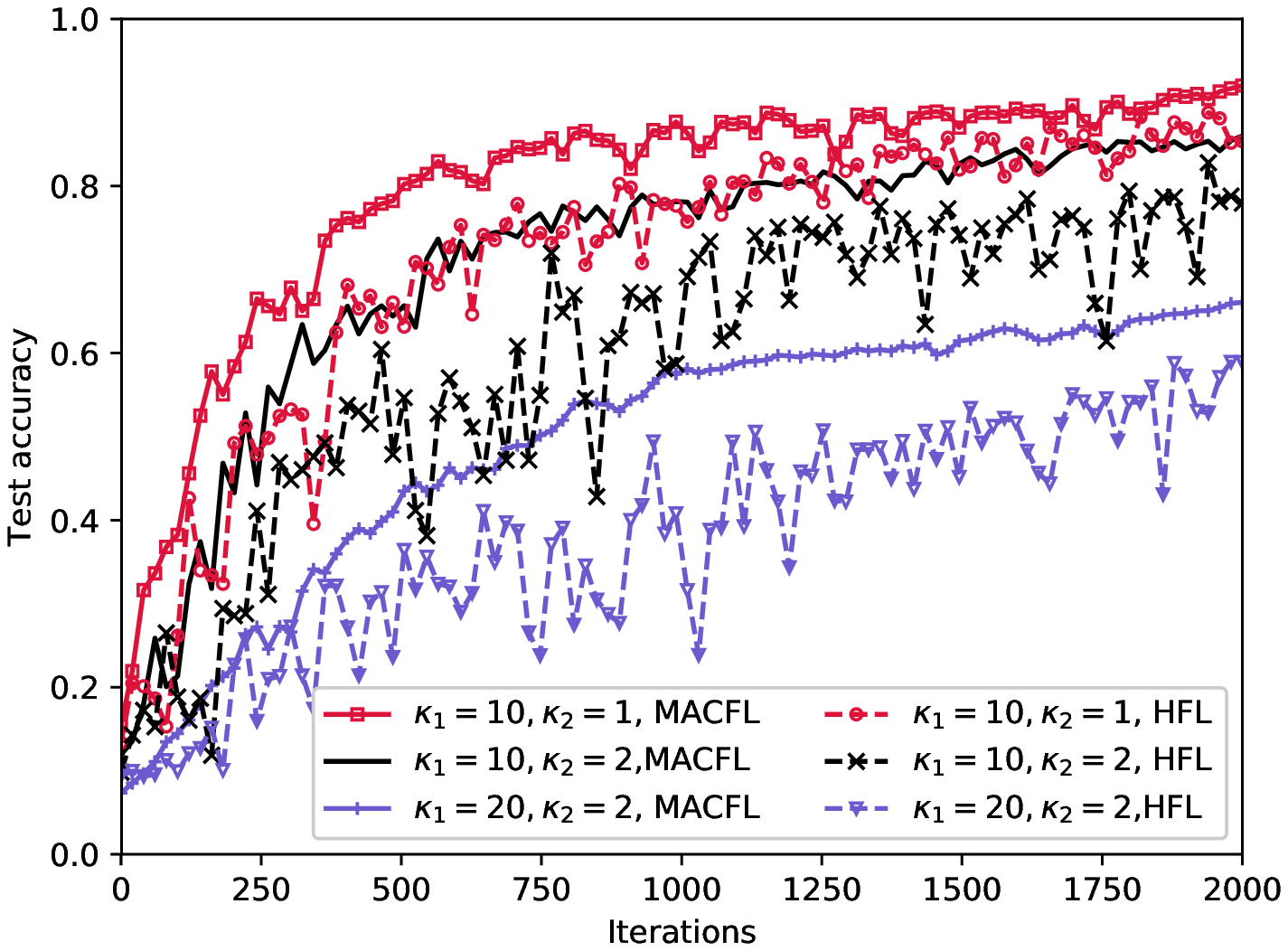}}
      \caption{Comparison of the evaluation of test accuracy performance between HFL and our proposed MACFL algorithm with different data distributions and system parameters $\kappa_1,\kappa_2$ when $p_s=0.5, M=50$ and $N=5$.}
      \label{fig:6}
\end{figure}

Fig. \ref{fig:6} plots the convergence rates of MACFL and HFL algorithm under different values of $\kappa_1$ and $\kappa_2$. Firstly, the convergence rate of our proposed MACFL algorithm can be enhanced with increasing $\kappa_1$ and $\kappa_2$ for both IID and non-IID data distributions, which are in consistence with Theorem ~\ref{2theorem}. 
Secondly, our proposed MACFL algorithm attains faster convergence rate than the HFL algorithm in all cases. 
% \textbf{These results substantiate the converge analysis given in Theorem ~\ref{2theorem}, and similar to HFL algorithm, more aggregations is useful for a performance improvement with MACFL algorithm.}
Particularly, for the IID dataset case, the test accuracy can be improved from $87.75\%$ to $89.10\%$ with $\kappa_1 = 20, \kappa_2 = 2$ by using our proposed scheme, and from $92.96\%$ to $94.01\%$ with  $\kappa_1 = 20, \kappa_2 = 1$, and from $93.32\%$ to $94.57\%$ with $\kappa_1 = 10, \kappa_2 = 1$. 
As for the non-IID datasetcase, the performance gains are more pronounced, whereas the test accuracy can be improved from $58.92\%$ to $66.07\%$ with $\kappa_1 = 20, \kappa_2 = 2$ with MACFL algorithm, and from $77.95\%$ to $86.01\%$ with  $\kappa_1 = 20, \kappa_2 = 1$, and from $85.35\%$ to $91.97\%$ with $\kappa_1 = 10, \kappa_2 = 1$. 
Moreover, the fluctuations in the convergence curve can also be largely reduced, especially for non-IID dataset case. 
Such an improvement in the learning performance is ascribed to the sophisticated weighted average scheme and more effective model update and aggregation rules employed in our proposed MACFL algorithm.

\begin{figure}
    \centering
     \subfigure[]{
  \includegraphics[width=0.45\textwidth]{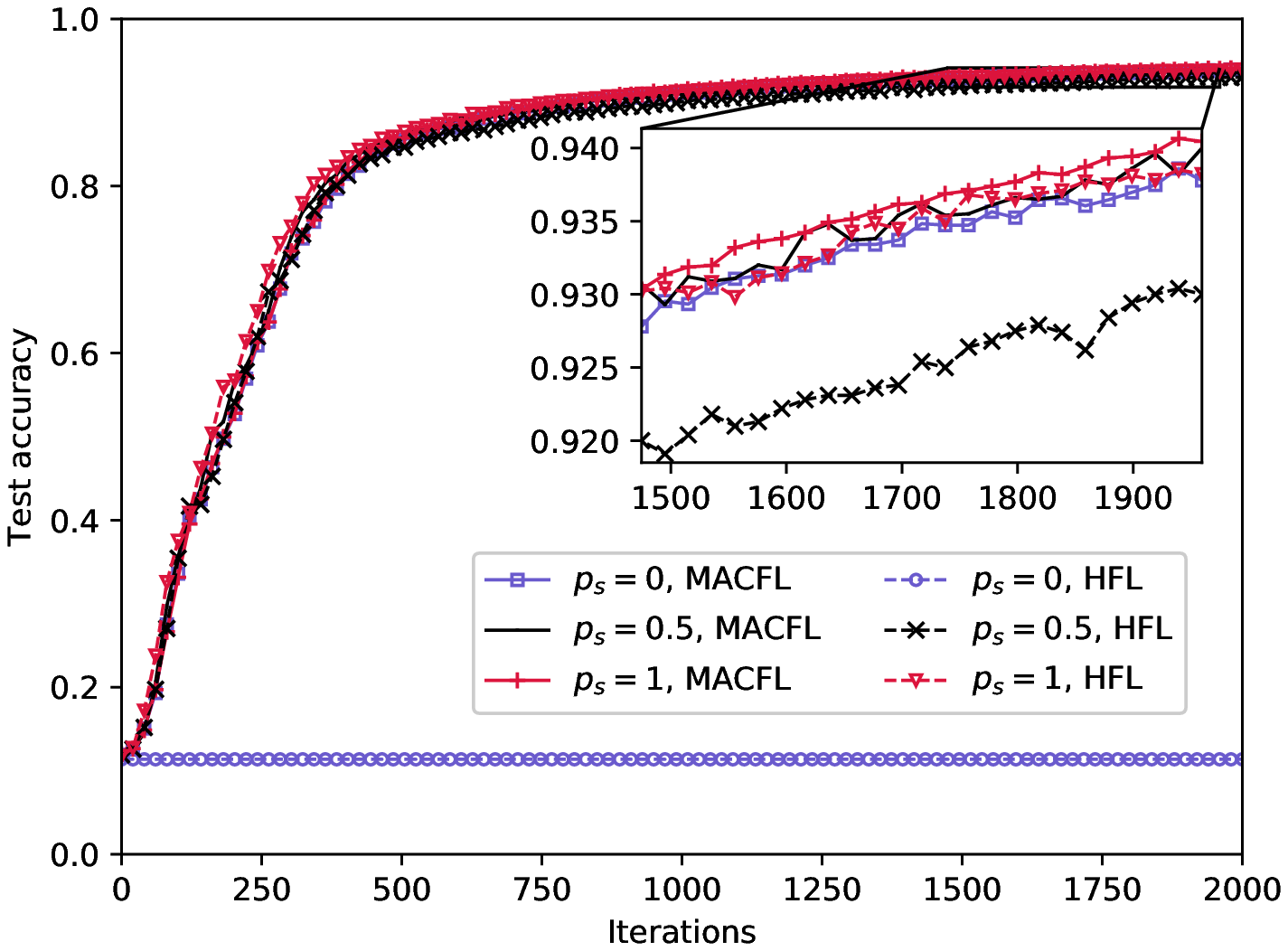} }
    \subfigure[]{
  \includegraphics[width=0.45\textwidth]{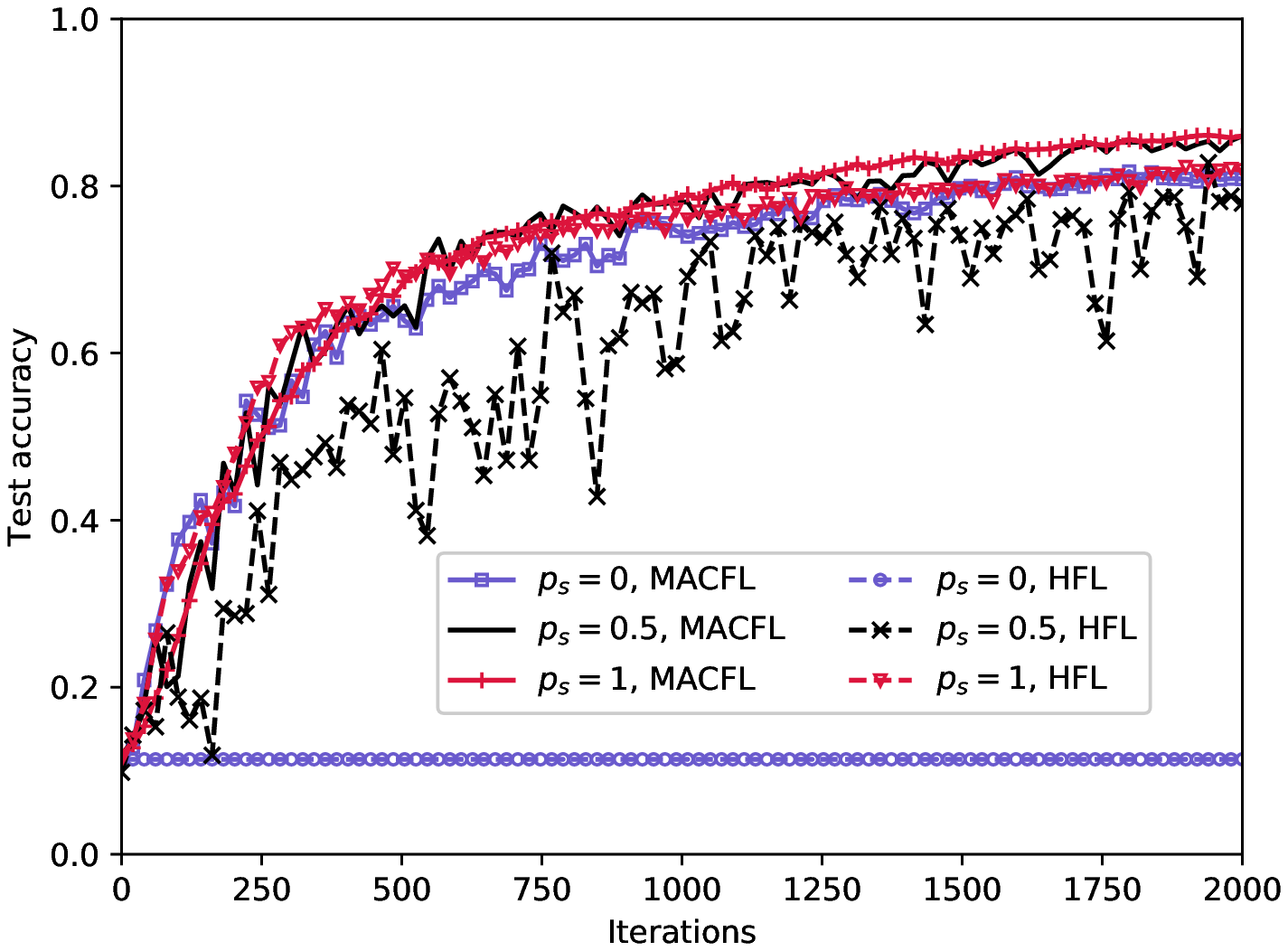} }
    \caption{Comparison of the evaluation of test accuracy performance between HFL and our proposed MACFL algorithm with different data distributions and system parameters $p_s$ when $ \kappa_1=20,\kappa_2=1, M=50$ and $N=5$ . }
    \label{fig:7}
\end{figure}

In Fig. \ref{fig:7}, we compare the performance of our proposed MACFL and conventional HFL algorithm under various staying probability $p_s$ of mobile users. 
The first noteworthy observation is that different from HFL algorithm, the test accuracy of our proposed MACFL algorithm is not sensitive to the variants of $p_s$ for both IID and non-IID data distributions. 
This phenomenon can be explained by our proposed model aggregation rule, in which the stay probability $p_s$ does not affect the number of effective participants.
Secondly, it shows that our proposed algorithm outperforms the conventional HFL algorithm with mobile users for different $p_s$ settings, especially for $p_s=0$, where the test accuracy can be improved from $11.37\%$ to $93.86\%$ for IID dataset case by using our proposed scheme, and from $11.37\%$ to $80.85\%$ for non-IID dataset case. 
Unlike HFL algorithm with $p_s=0$, in which the global model never get updated due to the extremely high mobility of users, our proposed model aggregation rule allows all the mobile user to upload their computed results at each communication round even though no user is connected with the same tagged edge AP during local training phase.
Moreover, our algorithm also helps to reduce the vibrations of convergence curves, especially for non-IID dataset case, owing to  attentive weighted scheme in the proposed algorithm.

 \begin{figure}
    \centering    
       \subfigure[]{
  \includegraphics[width=0.45\textwidth]{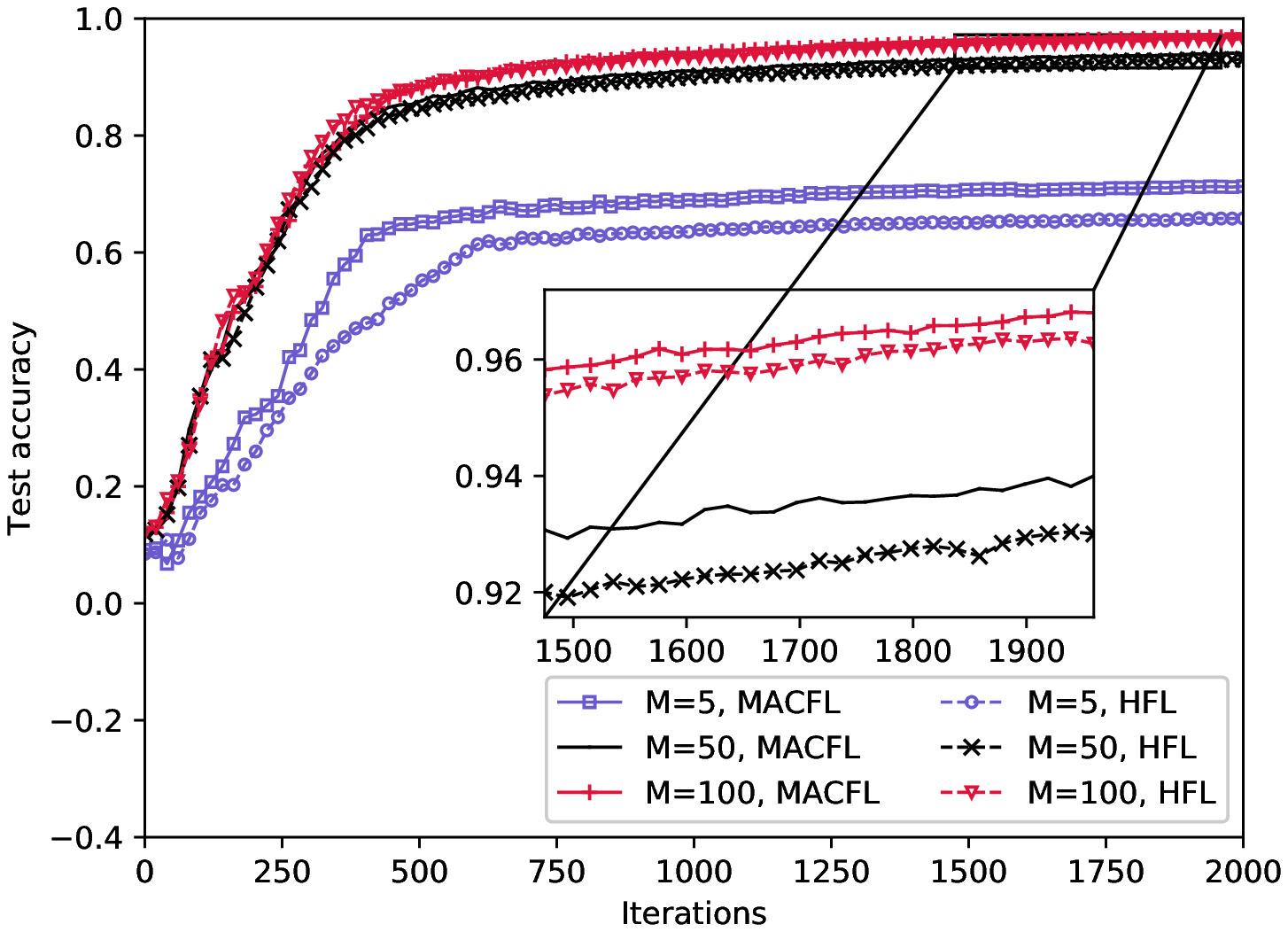} }
    \subfigure[]{
  \includegraphics[width=0.45\textwidth]{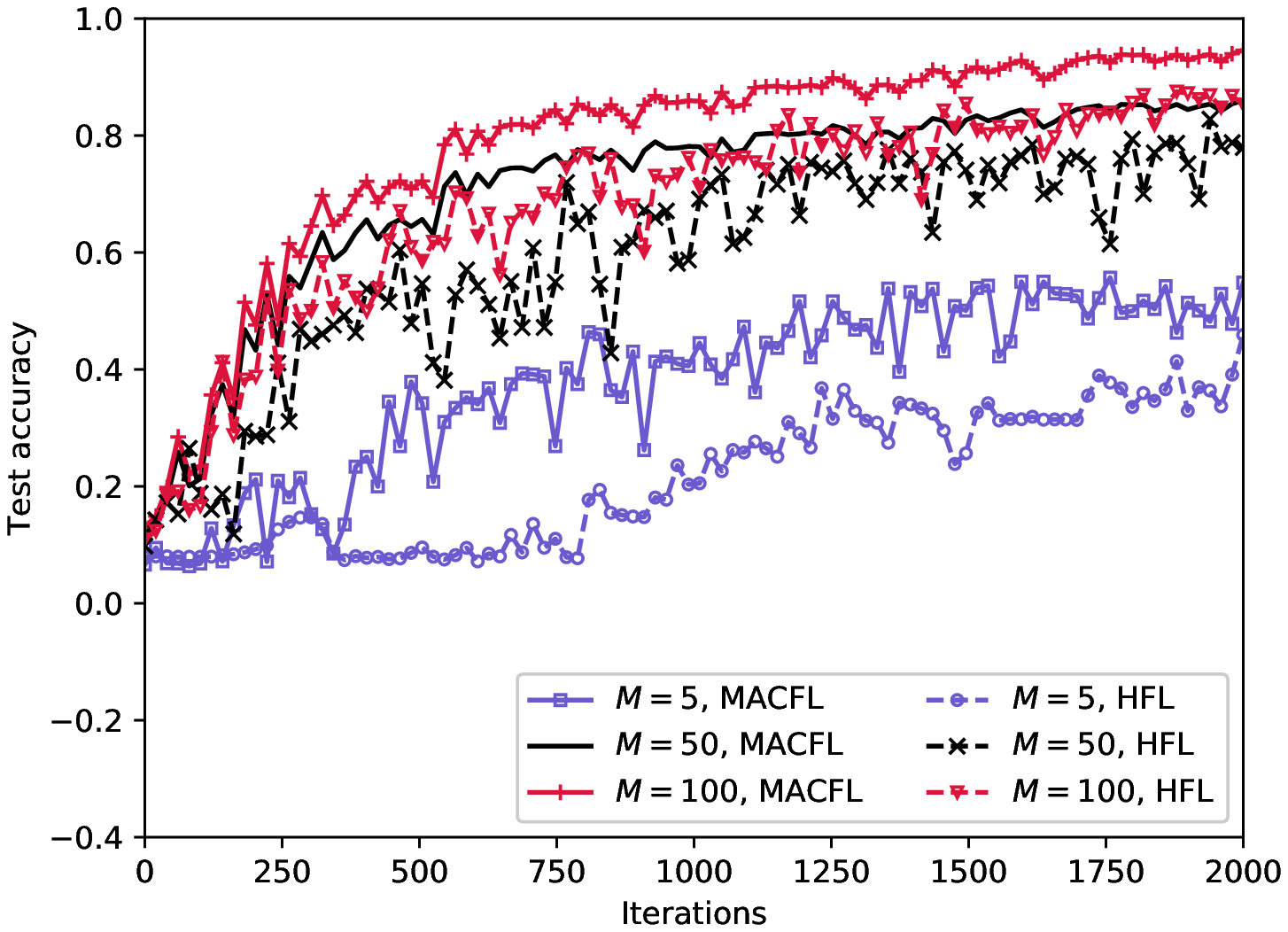} }
    \caption{Comparison of the evaluation of test accuracy performance between HFL and our proposed MACFL algorithm with different data distributions and system parameters $M$ when $p_s=0.5, \kappa_1=20,\kappa_2=1$ and $N=5$. }
    \label{fig:8}
\end{figure}

In Fig. \ref{fig:8}, the test accuracy performance of our proposed MACFL algorithm under different $M$ is evaluated, and conventional HFL with mobile users is chosen as a benchmark. 
Firstly, similar to HFL algorithm, we note that the test accuracy of our proposed MACFL algorithm can be improved by increasing $M$ for both IID and non-IID data distributions. 
This can be explained by the same reasons for HFL algorithm, and the results prove the correctness of our convergence analysis given in Theorem ~\ref{2theorem}.
Secondly,  it shows that our proposed algorithm outperforms the conventional HFL algorithm with mobile users, especially for small value of $M$.  
In Fig. \ref{fig:8}, it shows that for the IID dataset case, the test accuracy can be improved from $65.83\%$ to $71.31\%$ with $M=5$ with our proposed scheme, 
and from $93.06\%$ to $94.01\%$ with  $M=50$, 
and from $95.47\%$ to $95.93\%$ with $M=100$. 
As for non-IID dataset case, the improvements are more pronounced. In particular, as Fig. \ref{fig:8} shows, the test accuracy can be improved 
from $45.89\%$ to $56.85\%$ with $M=5$ by using our proposed averaging scheme, and from$77.95\%$ to $86.01\%$  with  $M=50$, and 
from $85.32\%$ to $93.60\%$ with $M=100$. 
Such enhancements mainly owes to our proposed model update and aggregation rules, which can help each user to learn a more accurate personalized local models and help both edge servers and cloud server to construct more precise shared models, and it also unveils the importance of a well designed averaging scheme in FL system.

% ============================================ %
%         Section: Conclusion                  %
% ============================================ %
\section{Conclusion}
\label{sec:conclusion}
In this paper, we studied the performance of FL in the context of a hierarchical wireless network consisted of one cloud server, multiple edge APs, and a large number of mobile users. 
We derived analytical expressions for the convergence rate of FL in the conventional setup, by accounting for several key factors of the system, including the heterogeneity arises from both the dataset and network architecture, as well as the user mobility. The analysis revealed that increasing the communication frequency amongst the users and their connected APs can accelerate the convergence rate. In contrast, an increase in user mobility leads to a dropout of participants and decreases the convergence rate. Furthermore, by exploiting the correlations amongst the parameters of mobile users and edge APs, we proposed a mobility-aware scheme to aggregate the users' parameters. The efﬁcacy of the proposed approach is corroborated via extensive experiments, whereas the gain over the conventional training method is particularly pronounced when mobile users possess non-IID datasets. Future extensions of this work can focus on the more delicate design of HFL with mobile users, such as personalized local model updates, data-driven model aggregation schemes, etc.

%%%%%%%%%%%%%%%%%%%%%%%%%%%%%%%%%%%%%%%%%%%%%%%%%%%%
\begin{appendix}
\subsection{Proof of Lemma~\ref{lemma1}}\label{ap:lemma1}
In this section, we use $g(\mathbf{w}_{u_m}^t)$, $F(\mathbf{w}_{u_m}^t)$, $f_{c_i}(\mathbf{w}_{c_i}^t)$ and $f (\mathbf{w}_g^t)$ as the short version of mini-batch gradients, local, cluster and global loss functions in this section. Similar to \cite{HLSGD2021,MLSGD2021}, we use averaged global model parameters to analysis even though the they are not observed for each iteration, and the virtual global model parameters can be written as $\mathbf{\bar{w}}_g^{t+1}=\mathbf{\bar{w}}_g^{t}-\eta\sum_{m=1}^M \alpha_{u_m} g(\mathbf{w}_{u_m}^t)  \mathbb{I}_{u_m}^t$ to analysis, then we have 
\begin{small}
\begin{align} \label{eq:ap1}
 \mathbb{E}f(\mathbf{\bar{w}}^{t+1}_g) & = \mathbb{E}f\left(\mathbf{\bar{w}}^{t}_g- \sum_{m=1}^M  \alpha_{u_m} g(\mathbf{w}_{u_m}^t) \mathbb{I}_{u_m}^t \right)\nonumber\\
& \overset{a} \leq \mathbb{E} f(\mathbf{\bar{w}}^{t}_g)+\frac{\eta^2L}{2} \mathbb{E} \| \sum_{m=1}^M \alpha_{u_m} g(\mathbf{w}_{u_m}^t)\mathbb{I}_{u_m}^t \|^2 -\eta\mathbb{E}< \nabla f(\mathbf{\bar{w}}^{t}), \sum_{m=1}^M  g(\mathbf{w}_{u_m}^t) \mathbb{I}_{u_m}^t >
\end{align}
\end{small}
where step (a) holds because of smoothness of the loss functions.
% %%%%%%%%%% second term %%%%%%%%%%
For the second term of in \eqref{eq:ap1}, we have 
\begin{small}
\begin{align}\label{eq:ap3}
 \mathbb{E} \| \sum_{m=1}^M \alpha_{u_m}  g(\mathbf{w}_{u_m}^t)\mathbb{I}_{u_m}^t \|^2
 \overset{a} = & \mathbb{E} \| \sum_{m=1}^M \alpha_{u_m} \left(g(\mathbf{w}_{u_m}^t)\mathbb{I}_{u_m}^t -p_s\nabla F(\mathbf{w}_{u_m}^t) \right) \|^2  + \mathbb{E} \| p_s \sum_{m = 1}^M\ \alpha_{u_m} \nabla F(\mathbf{w}_{u_m}^t) \|^2\nonumber\\
 \overset{b} \leq & \mathbb{E} \| \sum_{m=1}^M \alpha_{u_m} \left(g(\mathbf{w}_{u_m}^t)\mathbb{I}_{u_m}^t -p_s\nabla F(\mathbf{w}_{u_m}^t) \right) \|^2  + p_s^2  \sum_{m = 1}^M\ \alpha_{u_m} \mathbb{E} \|  \nabla F(\mathbf{w}_{u_m}^t) \|^2
\end{align}
\end{small}
where the step (a) in \eqref{eq:ap3} holds because $\mathbb{E}\| \mathbf{x} \|^2 = \mathbb{E}\| \mathbf{x}-\mathbb{E}\mathbf{x} \|^2 -(\mathbb{E}\mathbf{x})^2$, step (b) in \eqref{eq:ap3} holds because $ \| \sum_{i=1}^N a_i x_i  \|^2  \leq \sum_{i=1}^N a_i \| x_i \|^2 $, with $ 0 \leq a_i \leq 1, \sum_{i=1}^N a_i =1$.
For the first term in inequality \eqref{eq:ap3}, we have 
\begin{small}
\begin{align}\label{eq:ap31}
& \mathbb{E} \| \sum_{m=1}^M \alpha_{u_m} \left(g(\mathbf{w}_{u_m}^t)\mathbb{I}_{u_m}^t -p_s\nabla F(\mathbf{w}_{u_m}^t) \right) \|^2 \nonumber\\
  = &  \mathbb{E} \bigg\| \sum_{m=1}^M \alpha_{u_m} \bigg( g(\mathbf{w}_{u_m}^t)\mathbb{I}_{u_m}^t -\nabla F(\mathbf{w}_{u_m}^t)  +  (1-p_s) \nabla F(\mathbf{w}_{u_m}^t) \bigg) \bigg \|^2   \nonumber\\
   = &  \sum_{m=1}^M \alpha_{u_m}^2 \mathbb{E} \bigg\| g(\mathbf{w}_{u_m}^t)\mathbb{I}_{u_m}^t -\nabla F(\mathbf{w}_{u_m}^t)  + (1-p_s) \nabla F (\mathbf{w}_{u_m}^t)  \bigg\|^2+ \sum_{m=1}^M \sum_{j=1,j\neq m}^M \alpha_{u_m}  \alpha_{u_j} \mathbb{E} \bigg\{< g(\mathbf{w}_{u_m}^t) \mathbb{I}_{u_m}^t \nonumber\\
&   -\nabla F(\mathbf{w}_{u_m}^t) + (1-p_s) \nabla F(\mathbf{w}_{u_m}^t), g (\mathbf{w}_{u_j}^t)\mathbb{I}_{u_j}^t -\nabla F(\mathbf{w}_{u_j}^t) + (1-p_s) \nabla F(\mathbf{w}_{u_j}^t)> \bigg\}
\end{align} 
\end{small}
For the first term in inequality \eqref{eq:ap31}, we have 
\begin{small}
\begin{align}\label{eq:ft31}
&  \sum_{m=1}^M \alpha_{u_m}^2 \mathbb{E} \bigg\| g(\mathbf{w}_{u_m}^t)\mathbb{I}_{u_m}^t -\nabla F(\mathbf{w}_{u_m}^t) + (1-p_s) \nabla F (\mathbf{w}_{u_m}^t)  \bigg\|^2 \nonumber\\
 = &  \sum_{m=1}^M \alpha_{u_m}^2 
\bigg( \mathbb{E} \| g(\mathbf{w}_{u_m}^t)\mathbb{I}_{u_m}^t -\nabla F(\mathbf{w}_{u_m}^t)\|^2 +   (1-p_s)^2  \mathbb{E} \|  \nabla F(\mathbf{w}_{u_m}^t)  \|^2 +   \mathbb{E}  <  (1-p_s) \nabla F(\mathbf{w}_{u_m}^t),\nonumber\\ 
%& +  \sum_{m=1}^M \alpha_{u_m}^2  \bigg( \mathbb{E}  <  (1-p_s) \nabla F(\mathbf{w}_{u_m}^t),  g(\mathbf{w}_{u_m}^t)\mathbb{I}_{u_m}^t -\nabla F(\mathbf{w}_{u_m}^t)> \nonumber\\
 &  g(\mathbf{w}_{u_m}^t)\mathbb{I}_{u_m}^t -\nabla F(\mathbf{w}_{u_m}^t)> + \mathbb{E}  < g(\mathbf{w}_{u_m}^t)\mathbb{I}_{u_m}^t -\nabla F(\mathbf{w}_{u_m}^t), (1-p_s) \nabla F(\mathbf{w}_{u_m}^t) > \bigg) \nonumber\\
\overset{a} = & \sum_{m=1}^M \alpha_{u_m}^2 \mathbb{E}  \| g(\mathbf{w}_{u_m}^t)\mathbb{I}_{u_m}^t -\nabla F(\mathbf{w}_{u_m}^t)\|^2   -\sum_{m=1}^M \alpha_{u_m}^2  (1-p_s)^2    \mathbb{E} \|  \nabla F(\mathbf{w}_{u_m}^t) \|^2   \nonumber\\
\leq &\sum_{m=1}^M \alpha_{u_m}^2 \bigg( p_s \sigma^2 +p_s(1-p_s) \mathbb{E} \| \nabla F(\mathbf{w}_{u_m}^t) \|^2\bigg), 
\end{align}
\end{small}
where step (a) holds because the mini-batch gradients are assumed to be unbiased with bounded variance, and the independence between the mobility variable and the mini-batch sampling, we have $\mathbb{E}\{ g(\mathbf{w}_{u_m}^t )\mathbb{I}_{u_m}^t  \} = p_s \nabla F(\mathbf{w}_{u_m}^t)$. 
For the second term in inequality \eqref{eq:ap31}, we have 
%\begin{align}\label{eq:st31}& 
$\sum_{m=1}^M \sum_{j=1,j\neq m}^M \alpha_{u_m}  \alpha_{u_j} \mathbb{E} \big\{ < g(\mathbf{w}_{u_m}^t) \mathbb{I}_{u_m}^t -\nabla F(\mathbf{w}_{u_m}^t) + (1-p_s) \nabla F(\mathbf{w}_{u_m}^t), g (\mathbf{w}_{u_j}^t)\mathbb{I}_{u_j}^t -\nabla F(\mathbf{w}_{u_j}^t)$
$ + (1-p_s) \nabla F(\mathbf{w}_{u_j}^t)> \big\} =0 $ because of the unbiasedness of mini-batch gradients and the independence between user mobility and random sampling. 

Plugging $\eqref{eq:ft31}$ back into inequality \eqref{eq:ap3}, we have 
\begin{small}
\begin{align}\label{eq:ap32}
\mathbb{E} \|  \sum_{m=1}^M \alpha_{u_m}  g(\mathbf{w}_{u_m}^t)\mathbb{I}_{u_m}^t  \|^2  \leq p_s \sigma^2 \sum_{m=1}^M \alpha_{u_m}^2 +\sum_{m=1}^M \alpha_{u_m} p_s\left(p_s+ \alpha_{u_m} (1-p_s)\right)  \mathbb{E} \| \nabla F(\mathbf{w}_{u_m}^t) \|^2
\end{align}
\end{small}
For the third term in \eqref{eq:ap1}, we have 
\begin{small}
\begin{align}\label{eq:ap2}
& -\eta\mathbb{E} < \nabla f(\mathbf{\bar{w}}_g^{t}),  \sum_{m=1}^M \alpha_{u_m}  g_{u_m}(\mathbf{w}_{u_m}^t )\mathbb{I}_{u_m}^t> \nonumber\\
\overset{a} =&  - \eta p_s   \sum_{m=1}^M \alpha_{u_m}   \mathbb{E} < \nabla f(\mathbf{\bar{w}}_g^{t}), \nabla F(\mathbf{w}_{u_m}^t) >\nonumber\\
 \overset{b} = &  -\frac{\eta p_s}{2} \sum_{m=1}^M \alpha_{u_m}  \bigg( \mathbb{E} \| \nabla f(\mathbf{\bar{w}}^{t}_g)\|^2+\mathbb{E} \| \nabla F(\mathbf{w}_{u_m}^t) \|^2 -\mathbb{E} \| \nabla f(\mathbf{\bar{w}}^{t}_g)- \nabla F(\mathbf{w}_{u_m}^t) \|^2 \bigg)
\end{align}
\end{small}
where a conditional expectation is made in step (a), and step (b) holds because $<\mathbf{x},\mathbf{y}>= \frac{1}{2} ( \| \mathbf{x}-\mathbf{y} \|^2 - \| \mathbf{x} \|^2 - \| \mathbf{y} \|^2) $.
For the last term in \eqref{eq:ap2}, we have
\begin{small}
\begin{align}\label{eq:ap6}
& \sum_{m=1}^M \alpha_{u_m} \mathbb{E} \| \nabla f(\mathbf{\bar{w}}^{t}_g)- \nabla F(\mathbf{w}_{u_m}^t) \|^2 \nonumber\\
= & \sum_{m=1}^M \alpha_{u_m} \mathbb{E} \bigg\| \nabla f(\mathbf{\bar{w}}^{t}_g)\pm \nabla f(\mathbf{\bar{w}}^{t}_{c_{u_m}}) \pm \nabla f_{c_{u_m}}(\mathbf{\bar{w}}^{t}_{c_{u_m}}) \pm \nabla F(\mathbf{\bar{w}}^{t}_{c_{u_m}}) -  \nabla F(\mathbf{w}_{u_m}^t) \bigg\|^2 \nonumber\\
\overset{a} \leq & 4 L^2 \sum_{i=1}^N \alpha_{c_i} \mathbb{E} \| \mathbf{\bar{w}}^{t}_g - \mathbf{\bar{w}}_{c_i}^t \|^2 + 4\epsilon_g^2  +  4 L^2 \sum_{m=1}^M \alpha_{u_m}  \mathbb{E} \| \mathbf{\bar{w}}_{c_{u_m}}^t - \mathbf{w}_{u_m}^t \|^2 + 4 \epsilon_c^2 
\end{align}
\end{small}
where $\mathbf{\bar{w}}_{c_{u_m}}^t$ denotes the averaged cluster model owned by the edge AP which the $u_m$ located in at $t$-th iteration, step (a) holds because of the assumptions of smoothness and bounded divergences among the local, cluster and global loss functions.  

Plugging  \eqref{eq:ap32}, \eqref{eq:ap2} and \eqref{eq:ap6} back into inequality \eqref{eq:ap1}, and rearranging the order, dividing both side by $\frac{\eta p_s}{2}$ and taking the average over time, we have 
\begin{small}
\begin{align}\label{eq:ap5}
  \frac{1}{T} \sum_{t=1}^T \mathbb{E} \| \nabla f(\mathbf{\bar{w}}^{t}_g)\|^2  %\nonumber\\
&  \leq  \frac{2}{\eta p_s T}  (\mathbb{E} f (\mathbf{\bar{w}}^0_g) -f_{\inf})  + \eta L  \sigma^2 \sum_{m=1}^M \alpha_{u_m}^2   +  4\epsilon_g^2 + 4 \epsilon_c^2\nonumber\\ 
 &+  \frac{4 L^2}{T}\sum_{t=0}^{T-1}  \sum_{i=1}^N \alpha_{c_i} \mathbb{E} \| \mathbf{\bar{w}}^{t}_g - \mathbf{\bar{w}}_{c_i}^t \|^2 + \frac{4 L^2}{T}\sum_{t=0}^{T-1}  \sum_{m=1}^M \alpha_{u_m}  \mathbb{E} \| \mathbf{\bar{w}}_{c_{u_m}}^t - \mathbf{w}_{u_m}^t \|^2  \nonumber\\
   & +  \frac{1}{T}\sum_{t=0}^{T-1} \sum_{m=1}^M \alpha_{u_m} \left (\eta L(p_s+\alpha_{u_m}(1-p_s) )-1\right)   \mathbb{E} \| \nabla F(\mathbf{w}_{u_m}^t) \|^2
\end{align}
\end{small}
The last term in \eqref{eq:ap5} characterizes the efficiency of gradient descent with specific loss functions, if the learning rate is small enough, the last term is less than zero. Specifically, if  $\eta \leq \frac{1}{L}$, we have $\eta L(p_s+\alpha_{u_m} (1-p_s) ) \leq  1$ since $0 \leq p_s \leq 1$ and $ 0 \leq \alpha_{u_m} \leq 1$. By far, we conclude the proof of  Lemma \ref{lemma1}.

\subsection{Proof of Lemma \ref{lemma2}}\label{ap:lemma2}
%%%%% local mse 
For the MSE of the local and cluster model parameters, if $t \mod \kappa_1 =0$, we have $\mathbf{w}_{u_m}^t=\mathbf{w}_{c_{u_m}}^t$ ,  otherwise, let $b= \lfloor \frac{t}{\kappa_1} \rfloor$, we have
\begin{small}
\begin{align}\label{eq:lem21}
& \sum_{m=1}^M  \alpha_{u_m} \mathbb{E} \| \mathbf{w}^{t}_{u_m}- \mathbf{\bar{w}}_{c_{u_m}}^t \|^2 \nonumber\\
=&\sum_{m=1}^M  \alpha_{u_m} \mathbb{E} \bigg \| (\mathbf{w}^{b\kappa_1}_{u_m}- \eta\sum_{\tau=b\kappa_1}^{t-1} g(\mathbf{w}^{\tau}_{u_m})\mathbb{I}^{\tau}_{u_m} - (\mathbf{\bar{w}}_{c_{u_m}}^{b\kappa_1} -\eta\sum_{u_m\in \mathcal{C}_{u_m}} \sum_{\tau=b\kappa_1}^{t-1}  \alpha_{u_m}^c g(\mathbf{w}^{\tau}_{u_m}) \mathbb{I}^{\tau}_{u_m})\bigg \|^2 \nonumber\\
=& \eta^2  \sum_{m=1}^M  \alpha_{u_m} \mathbb{E} \bigg\| \sum_{\tau=b\kappa_1}^{t-1} \bigg( \sum_{u_m\in \mathcal{C}_{u_m}} \alpha_{u_m}^c g(\mathbf{w}^{\tau}_{u_m}) \mathbb{I}^{\tau}_{u_m}-g(\mathbf{w}^{\tau}_{u_m})\mathbb{I}^{\tau}_{u_m} \bigg)\bigg\|^2 \nonumber\\
=&\eta^2  \sum_{m=1}^M  \alpha_{u_m} \mathbb{E} \bigg\| \sum_{\tau=b\kappa_1}^{t-1} \bigg( \sum_{u_m\in \mathcal{C}_{u_m}} \alpha_{u_m}^c g(\mathbf{w}^{\tau}_{u_m}) \mathbb{I}^{\tau}_{u_m}
\pm p_s \sum_{u_m \in \mathcal{C}_{u_m} }\alpha_{u_m}^c \nabla F(\mathbf{w}^{\tau}_{u_m}) \pm p_s \nabla F(\mathbf{w}^{\tau}_{u_m}) %\nonumber\\ &
-g(\mathbf{w}^{\tau}_{u_m})\mathbb{I}^{\tau}_{u_m} \bigg)\bigg\|^2 \nonumber\\
\leq & 2\eta^2  \sum_{m=1}^M  \alpha_{u_m} \bigg\{ \mathbb{E} \bigg\| \sum_{\tau=b\kappa_1}^{t-1} \bigg[ 
\sum_{u_m\in \mathcal{C}_{u_m}} \alpha_{u_m}^c \bigg( g(\mathbf{w}^{\tau}_{u_m}) \mathbb{I}^{\tau}_{u_m}-  p_s \nabla F(\mathbf{w}^{\tau}_{u_m})\bigg)
-\bigg( g(\mathbf{w}^{\tau}_{u_m}) \mathbb{I}^{\tau}_{u_m}- p_s \nabla F(\mathbf{w}^{\tau}_{u_m}) \bigg)  \bigg] \bigg\|^2 \nonumber\\
&
+ p_s^2 \mathbb{E} \bigg\| \sum_{\tau=b\kappa_1}^{t-1} \bigg(\sum_{u_m \in \mathcal{C}_{u_m} } \alpha_{u_m}^c \nabla F(\mathbf{w}^{\tau}_{u_m})- \nabla F(\mathbf{w}^{\tau}_{u_m}) \bigg) \bigg\|^2 \bigg\}
\end{align}
\end{small}
For the first term in \eqref{eq:lem21}, we have 
\begin{small}
\begin{align}\label{eq:lem22}
&   \sum_{m=1}^M  \alpha_{u_m} \bigg\{ \mathbb{E} \bigg\| \sum_{\tau=b\kappa_1}^{t-1} \bigg[ 
\sum_{u_m\in \mathcal{C}_{u_m}} \alpha_{u_m}^c \bigg( g(\mathbf{w}^{\tau}_{u_m}) \mathbb{I}^{\tau}_{u_m}- p_s \nabla F(\mathbf{w}^{\tau}_{u_m})\bigg)
-\bigg( g(\mathbf{w}^{\tau}_{u_m}) \mathbb{I}^{\tau}_{u_m}- p_s \nabla F(\mathbf{w}^{\tau}_{u_m}) \bigg)  \bigg] \bigg\|^2\nonumber\\
 \overset{a}= & \sum_{n=1}^N  \sum_{u_m \in \mathcal{C}_n} \alpha_{c_n} \alpha_{u_m}^c \bigg[\mathbb{E} \bigg\| \sum_{\tau=b\kappa_1}^{t-1} \bigg( g(\mathbf{w}^{\tau}_{u_m})\mathbb{I}^{\tau}_{u_m}  - p_s\nabla F(\mathbf{w}^{\tau}_{u_m})  - \sum_{u_m\in \mathcal{C}_{u_m}} \alpha_{u_m}^c (g(\mathbf{w}^{\tau}_{u_m}) \mathbb{I}^{\tau}_{u_m} %\nonumber\\ &
 - p_s \nabla F(\mathbf{w}^{\tau}_{u_m}) )\bigg)\bigg\|^2\bigg]\nonumber\\
  \overset{b}= & \sum_{m=1}^M \alpha_{u_m} \mathbb{E} \big\| \sum_{\tau=b\kappa_1}^{t-1} \big( g(\mathbf{w}^{\tau}_{u_m})\mathbb{I}^{\tau}_{u_m} - p_s \nabla F(\mathbf{w}^{\tau}_{u_m}) \big) \big\|^2  -  \sum_{n=1}^N  \alpha_{c_n} \mathbb{E} \big\| \sum_{\tau=b\kappa_1}^{t-1} \sum_{u_m \in \mathcal{C}_n}  \alpha_{u_m}^c   \big( g(\mathbf{w}^{\tau}_{u_m}) \mathbb{I}^{\tau}_{u_m}% \nonumber\\&
  - p_s\nabla F(\mathbf{w}^{\tau}_{u_m}) \big)\big\|^2 \nonumber\\
  \overset{c} = &    \sum_{m=1}^M  \sum_{\tau=b\kappa_1}^{t-1} \alpha_{u_m}   \bigg(\mathbb{E} \bigg\| g(\mathbf{w}^{\tau}_{u_m}) \mathbb{I}^{\tau}_{u_m}- p_s\nabla F(\mathbf{w}^{\tau}_{u_m})  \bigg\|^2  -    \alpha_{u_m}^c \mathbb{E} \bigg\|  g(\mathbf{w}^{\tau}_{u_m}) \mathbb{I}^{\tau}_{u_m} %\nonumber\\&
 - p_s\nabla F(\mathbf{w}^{\tau}_{u_m}) \bigg\|^2 \bigg)\nonumber\\
\leq &    \kappa_1 \sum_{m=1}^M \alpha_{u_m} (1-\alpha_{u_m}^c) \mathbb{E} \bigg\| g(\mathbf{w}^{\tau}_{u_m}) \mathbb{I}^{\tau}_{u_m}- p_s\nabla F(\mathbf{w}^{\tau}_{u_m})  \bigg\|^2 
\end{align}
\end{small}
where step (a) holds because $\alpha_{u_m} = \alpha_{u_m}^c \alpha_{c_{u_m}}$, step (b) holds because $\sum_{i=1}^N a_i \| x_i -\sum_{i=1}^N a_i x_i \|^2 = \sum_{i=1}^N a_i \| x_i \|^2  -(\sum_{i=1}^N a_i x_i)^2$ with $\sum_{i=1}^N a_i =1, 0 \leq  a_i \leq 1$,  step (c)  holds because $ \| \sum_{i=1}^N x_i -\mathbb{E} \{x_i\}  \|^2 = \sum_{i=1}^N \| x_i -\mathbb{E} \{x_i\} \|^2 $ with the independence among different $x_i$. And then we have $4\kappa_1 \sum_{m=1}^M \alpha_{u_m} (1-\alpha_{u_m}^c) \mathbb{E} \bigg\| g(\mathbf{w}^{\tau}_{u_m}) \mathbb{I}^{\tau}_{u_m}- p_s\nabla F(\mathbf{w}^{\tau}_{u_m})  \bigg\|^2 \leq   \kappa_1p_s (\sigma^2+(1-p_s) G^2) \sum_{m=1}^M \alpha_{u_m} (1-\alpha_{u_m}^c)$ because  $\mathbb{E} \| x_i -\mathbb{E} \{x_i\} \|^2 = \mathbb{E} \| x_i \|^2  -(\mathbb{E} \{x_i\})^2$.

For the second term in \eqref{eq:lem21}, we have 
\begin{small}
\begin{align}\label{eq:lem23}
& \sum_{m=1}^M \alpha_{u_m} \mathbb{E} \| \sum_{\tau=b\kappa_1}^{t-1} (\nabla F(\mathbf{w}^{\tau}_{u_m}) -\sum_{u_m \in \mathcal{C}_{u_m} } \alpha_{u_m}^c \nabla F(\mathbf{w}^{\tau}_{u_m}) ) \|^2 \nonumber\\
= &  \sum_{m=1}^M \alpha_{u_m} \mathbb{E} \bigg\| \sum_{\tau=b\kappa_1}^{t-1} \bigg(\nabla F(\mathbf{w}^{\tau}_{u_m}) \pm \nabla F(\mathbf{\bar{w}}^{\tau}_{c_{u_m}}) \pm \nabla f_{c_{u_m}} (\mathbf{\bar{w}}^{\tau}_{c_{u_m}}) -\sum_{u_j \in \mathcal{C}_{u_m} } \alpha_{u_m}^c \nabla F(\mathbf{w}^{\tau}_{u_j})\bigg)\bigg\|^2 \nonumber\\
 \leq & 3  \sum_{m=1}^M \alpha_{u_m} \bigg(
\mathbb{E} \bigg\| \sum_{\tau=b\kappa_1}^{t-1} \bigg(\nabla F(\mathbf{w}^{\tau}_{u_m}) - \nabla F(\mathbf{\bar{w}}^{\tau}_{c_{u_m}})\bigg) \bigg\|^2 + \mathbb{E} \bigg\|  \sum_{\tau=b\kappa_1}^{t-1} \bigg( \nabla F(\mathbf{\bar{w}}^{\tau}_{c_{u_m}})- \nabla f_{c_{u_m}}(\mathbf{\bar{w}}^{\tau}_{c_{u_m}}) \bigg) \bigg\|^2  \nonumber\\
&  +  \mathbb{E} \bigg\|  \sum_{\tau=b\kappa_1}^{t-1} ( \nabla f_{c_{u_m}} (\mathbf{\bar{w}}^{\tau}_{c_{u_m}}) -\sum_{u_j \in \mathcal{C}_{u_m} } \alpha_{u_m}^c \nabla F(\mathbf{w}^{\tau}_{u_j}) )\bigg\|^2 \bigg)\nonumber\\
 \leq  &  6 L^2\kappa_1   \sum_{m=1}^M \sum_{\tau=b\kappa_1}^{t-1} \alpha_{u_m}
\mathbb{E} \| \mathbf{w}^{\tau}_{u_m} - \mathbf{\bar{w}}^{\tau}_{c_{u_m}} \|^2 +  3 \kappa_1^2 \epsilon_c^2   
\end{align}
\end{small}
Plugging \eqref{eq:lem22} and \eqref{eq:lem23} back into \eqref{eq:lem23}, and taking average of time, if $\eta < \frac{1}{\sqrt{12}L\kappa_1}$ , we can conclude the proof of Lemma \ref{lemma2}. 
 
 \subsection{Proof of Lemma \ref{lemma3}}\label{ap:lemma3}
 %%%%%%%%%%%clustser mse
Similar to \cite{HLSGD2021} \cite{MLSGD2021}, we use averaged cluster model parameters to analysis even though the they are not observed for each iteration, and the virtual cluster model parameters can be written as $\mathbf{\bar{w}}_{c_n}^{t+1}=\mathbf{\bar{w}}_{c_n}^{t}-\eta \sum_{u_m \in \mathcal{C}_{n}^t } \alpha_{u_m}^c g(\mathbf{w}_{u_m}^t)  \mathbb{I}_{u_m}^t$ to analysis. We use $c(u_m)$ to indicate the index of edge AP that is connected by mobile user $u_m$ in the rest of this paper. For the MSE of the cluster and global model parameters, if $t \mod \kappa_1\kappa_2 =0$, we have $\mathbf{w}_{c_n}^t=\mathbf{w}_g^t$ if $t \mod \kappa_1\kappa_2 =0$,  otherwise, let $b_1= \lfloor \frac{t}{\kappa_1\kappa_2} \rfloor$. Similar to Lemma \ref{lemma2}, we have
\begin{small}
\begin{align}\label{eq:lem31}
&\sum_{n=1}^N \alpha_{c_n}  \mathbb{E} \| \mathbf{\bar{w}}^{t}_g - \mathbf{\bar{w}}_{c_n}^t \|^2 %\nonumber\\
%= &  \sum_{n=1}^N \alpha_{c_n}  \mathbb{E} \bigg\| (\mathbf{\bar{w}}^{b_1\kappa_1\kappa_2}_{g}- \eta \sum_{\tau=b_1\kappa_1\kappa_2}^{t-1} \sum_{m=1}^M \alpha_{u_m} g(\mathbf{w}^{\tau}_{u_m}) \mathbb{I}^{\tau}_{u_m}) - (\mathbf{\bar{w}}_{c_{u_m}}^{b_1\kappa_1\kappa_2} -\eta  \sum_{\tau=b_1\kappa_1\kappa_2}^{t-1} \sum_{u_m \in \mathcal{C}_{u_m}^{\tau}} \alpha_{u_m}^c g(\mathbf{w}^{\tau}_{u_m}) \mathbb{I}^{\tau}_{u_m})\bigg\|^2 \nonumber\\
%= & \eta^2  \sum_{n=1}^N \alpha_{c_n}  \mathbb{E} \bigg\| \sum_{\tau=b_1\kappa_1\kappa_2}^{t-1}  \bigg( \sum_{m=1}^M \alpha_{u_m} g(\mathbf{w}^{\tau}_{u_m}) \mathbb{I}^{\tau}_{u_m} -\sum_{u_m\in \mathcal{C}_{u_m}^{\tau} } \alpha_{u_m}^c g(\mathbf{w}^{\tau}_{u_m}) \mathbb{I}^{\tau}_{u_m}\bigg)\bigg\|^2 \nonumber\\
=  \eta^2  \sum_{n=1}^N  \alpha_{c_n} \mathbb{E} \bigg\| \sum_{\tau=b_1\kappa_1\kappa_2}^{t-1}  \bigg( \sum_{m=1}^M \alpha_{u_m} g(\mathbf{w}^{\tau}_{u_m}) \mathbb{I}^{\tau}_{u_m} \pm p_s  \nabla f_{c_{u_m}} (\mathbf{\bar{w}}^{\tau}_{c(u_m)})\nonumber\\
&  \pm p_s \sum_{j=1}^N \alpha_{c_j} \nabla f_{c_j} (\mathbf{\bar{w}}^{\tau}_{c_j}) -\sum_{u_m\in \mathcal{C}_{u_m}^{\tau} } \alpha_{u_m}^c g(\mathbf{w}^{\tau}_{u_m}) \mathbb{I}^{\tau}_{u_m}\bigg)\bigg\|^2 \nonumber\\
\leq &  2 \eta^2 \sum_{n=1}^N \alpha_{c_n}  \mathbb{E} \bigg\| \sum_{\tau=b_1\kappa_1\kappa_2}^{t-1}  \bigg( \sum_{u_m \in \mathcal{C}_{u_m}}  \alpha_{u_m}^cg(\mathbf{w}^{\tau}_{u_m}) \mathbb{I}^{\tau}_{u_m}  - p_s \nabla f_{c_{u_m}} (\mathbf{\bar{w}}^{\tau}_{c(u_m)}) -(\sum_{u_m\in \mathcal{C}_{u_m}} \alpha_{u_m}^c g(\mathbf{w}^{\tau}_{u_m}) \mathbb{I}^{\tau}_{u_m}\nonumber\\
& - p_s \sum_{j=1}^N \alpha_{c_j} \nabla f_{c_j} (\mathbf{\bar{w}}^{\tau}_{c_j})) \bigg)\bigg\|^2 +  2 \eta^2 p_s^2 \sum_{n=1}^N \alpha_{c_n} \mathbb{E} \bigg\| \sum_{\tau=b_1\kappa_1\kappa_2}^{t-1}  \bigg( \nabla f_{c_{u_m}} (\mathbf{\bar{w}}^{\tau}_{c(u_m)})- \sum_{j=1}^N \alpha_{c_j} \nabla f_{c_j} (\mathbf{\bar{w}}^{\tau}_{c_j})\bigg) \bigg\|^2 \nonumber\\
%\end{align}
%\end{small}
%For the first term in \eqref{eq:lem31}, we have 
%\begin{small}
%\begin{align}\label{eq:lem32}
%&  \sum_{n=1}^N \alpha_{c_n}  \mathbb{E} \bigg\| \sum_{\tau=b_1\kappa_1\kappa_2}^{t-1}  \bigg( \sum_{u_m \in \mathcal{C}_{u_m}}  \alpha_{u_m}^cg(\mathbf{w}^{\tau}_{u_m}) \mathbb{I}^{\tau}_{u_m}  - p_s \nabla f_{c_{u_m}} (\mathbf{\bar{w}}^{\tau}_{c(u_m)}) -(\sum_{u_m\in \mathcal{C}_{u_m}} \alpha_{u_m}^c g(\mathbf{w}^{\tau}_{u_m}) \mathbb{I}^{\tau}_{u_m}\nonumber\\
%& - p_s \sum_{j=1}^N \alpha_{c_j} \nabla f_{c_j} (\mathbf{\bar{w}}^{\tau}_{c_j})) \bigg)\bigg\|^2 \nonumber\\
%= & \sum_{n=1}^N \alpha_{c_n}  \bigg[ \mathbb{E} \bigg\| \sum_{\tau=b_1\kappa_1\kappa_2}^{t-1}  \bigg( ( \sum_{u_m \in \mathcal{C}_{u_m}^{\tau}}  \alpha_{u_m}^c (g(\mathbf{w}^{\tau}_{u_m}) \mathbb{I}^{\tau}_{u_m} \pm p_s \nabla F(\mathbf{w}^{\tau}_{u_m}) ) - p_s \nabla f_{c_{u_m}} (\mathbf{\bar{w}}^{\tau}_{c(u_m)})) \nonumber\\&-(\sum_{u_m\in \mathcal{C}_{u_m}^{\tau}} \alpha_{u_m}^c (g(\mathbf{w}^{\tau}_{u_m}) \mathbb{I}^{\tau}_{u_m}\pm p_s \sum_{m=1}^M \alpha_{u_m} \nabla F(\mathbf{w}^{\tau}_{u_m}) ) - p_s \sum_{j=1}^N \alpha_{c_j} \nabla f_{c_j} (\mathbf{\bar{w}}^{\tau}_{c_j})) \bigg)\bigg\|^2 \bigg]\nonumber\\
\leq & 4\eta^2 \sum_{n=1}^N \alpha_{c_n} \mathbb{E} \bigg\| \sum_{\tau=b_1\kappa_1\kappa_2}^{t-1}  \bigg(  \sum_{u_m \in \mathcal{C}_{u_m}^{\tau}}  \alpha_{u_m}^c (g(\mathbf{w}^{\tau}_{u_m}) \mathbb{I}^{\tau}_{u_m} -  p_s \nabla F(\mathbf{w}^{\tau}_{u_m}) ) - \sum_{m=1}^M \alpha_{u_m} (g(\mathbf{w}^{\tau}_{u_m}) \mathbb{I}^{\tau}_{u_m}   \nonumber\\
& - p_s \nabla F(\mathbf{w}^{\tau}_{u_m}) )\bigg)\bigg\|^2 + 4 \eta^2 p_s^2 \sum_{n=1}^N \alpha_{c_n}  \mathbb{E} \bigg\| \sum_{\tau=b_1\kappa_1\kappa_2}^{t-1}  \bigg( \sum_{u_m \in \mathcal{C}_{u_m}^{\tau}}  \alpha_{u_m}^c\nabla F(\mathbf{w}^{\tau}_{u_m}) - \nabla f_{c_{u_m}}(\mathbf{\bar{w}}^{\tau}_{c_{u_m}}) \bigg)\bigg\|^2\nonumber\\
&+  2 \eta^2 p_s^2 \sum_{n=1}^N \alpha_{c_n} \mathbb{E} \bigg\| \sum_{\tau=b_1\kappa_1\kappa_2}^{t-1}  \bigg( \nabla f_{c_{u_m}} (\mathbf{\bar{w}}^{\tau}_{c(u_m)})- \sum_{j=1}^N \alpha_{c_j} \nabla f_{c_j} (\mathbf{\bar{w}}^{\tau}_{c_j})\bigg) \bigg\|^2
\end{align}
\end{small}
For the first term in \eqref{eq:lem31}, we have 
\begin{small}
\begin{align}\label{eq:lem33}
& \sum_{n=1}^N \alpha_{c_n} \mathbb{E} \bigg\| \sum_{\tau=b_1\kappa_1\kappa_2}^{t-1}  \bigg(  \sum_{u_m \in \mathcal{C}_{u_m}^{\tau}}  \alpha_{u_m}^c (g(\mathbf{w}^{\tau}_{u_m}) \mathbb{I}^{\tau}_{u_m} -  \nabla F(\mathbf{w}^{\tau}_{u_m}) ) - p_s \sum_{m=1}^M \alpha_{u_m} (g(\mathbf{w}^{\tau}_{u_m}) \mathbb{I}^{\tau}_{u_m} %\nonumber\\& 
- p_s \nabla F(\mathbf{w}^{\tau}_{u_m}) )\bigg)\bigg\|^2  \nonumber\\
\overset{a}  = &  \sum_{n=1}^N \alpha_{c_n} 
\mathbb{E} \bigg\| \sum_{\tau=b_1\kappa_1\kappa_2}^{t-1} \bigg( \sum_{u_m \in \mathcal{C}_{u_m}}  \alpha_{u_m}^c  (g(\mathbf{w}^{\tau}_{u_m}) \mathbb{I}^{\tau}_{u_m}  -   p_s \nabla F(\mathbf{w}^{\tau}_{u_m}) ) \bigg) \bigg\|^2 - \mathbb{E} \bigg\| \sum_{\tau=b_1\kappa_1\kappa_2}^{t-1} \bigg( 
 \sum_{m=1}^M \alpha_{u_m} (g(\mathbf{w}^{\tau}_{u_m}) \mathbb{I}^{\tau}_{u_m}\nonumber\\
 &    - p_s \nabla F(\mathbf{w}^{\tau}_{u_m}) )\bigg)\bigg\|^2 \nonumber\\
%= &  \sum_{\tau=b_1\kappa_1\kappa_2}^{t-1} \sum_{n=1}^N \alpha_{c_n} \sum_{u_m \in \mathcal{C}_{u_m}}  \alpha_{u_m}^c   \alpha_{u_m}^c  \mathbb{E} \|g(\mathbf{w}^{\tau}_{u_m}) \mathbb{I}^{\tau}_{u_m} - p_s \nabla F(\mathbf{w}^{\tau}_{u_m})\|^2 - \sum_{\tau=b_1\kappa_1\kappa_2}^{t-1} \sum_{m=1}^M  \alpha_{u_m}^2 \nonumber\\
%&  \times \mathbb{E} \|g(\mathbf{w}^{\tau}_{u_m}) \mathbb{I}^{\tau}_{u_m}- p_s \nabla F(\mathbf{w}^{\tau}_{u_m})\|^2 \nonumber\\
= &  \sum_{\tau=b_1\kappa_1\kappa_2}^{t-1} \sum_{m=1}^M \alpha_{u_m} \bigg(  \alpha_{u_m}^c  \mathbb{E} \|g(\mathbf{w}^{\tau}_{u_m}) \mathbb{I}^{\tau}_{u_m} - p_s \nabla F(\mathbf{w}^{\tau}_{u_m})\|^2 - \alpha_{u_m} \mathbb{E} \|g(\mathbf{w}^{\tau}_{u_m}) \mathbb{I}^{\tau}_{u_m}- p_s \nabla F(\mathbf{w}^{\tau}_{u_m})\|^2 \bigg)\nonumber\\
%=&  \sum_{\tau=b_1\kappa_1\kappa_2}^{t-1} \sum_{m=1}^M \alpha_{u_m}(\alpha_{u_m}^c-\alpha_{u_m})  \mathbb{E} \|g(\mathbf{w}^{\tau}_{u_m}) \mathbb{I}^{\tau}_{u_m}- p_s \nabla F(\mathbf{w}^{\tau}_{u_m})\|^2 \nonumber\\
\overset{b} =&  \sum_{\tau=b_1\kappa_1\kappa_2}^{t-1} \sum_{m=1}^M \alpha_{u_m}(\alpha_{u_m}^c-\alpha_{u_m})  \bigg(\mathbb{E}\bigg\|  g(\mathbf{w}^{\tau}_{u_m}) \mathbb{I}^{\tau}_{u_m}- \nabla F(\mathbf{w}^{\tau}_{u_m})\bigg\|^2 - \mathbb{E} \bigg\| (p_s-1) \nabla F(\mathbf{w}^{\tau}_{u_m})\bigg\|^2\bigg)   \nonumber\\
%\overset{c} 
\leq  &  \kappa_1 \kappa_2 p_s ( \sigma^2 + (1-p_s)G^2  ) \sum_{m=1}^M \alpha_{u_m}(\alpha_{u_m}^c-\alpha_{u_m})
\end{align}
\end{small}
where step (a) holds because $\sum_{i=1}^N a_i \| x_i -\sum_{i=1}^N a_i x_i \|^2 = \sum_{i=1}^N a_i \| x_i \|^2  -(\sum_{i=1}^N a_i x_i)^2$ with $\sum_{i=1}^N a_i =1, 0 \leq  a_i \leq 1$, step (b) holds because  $\mathbb{E} \| x_i -\mathbb{E} \{x_i\} \|^2 = \mathbb{E} \| x_i \|^2  -(\mathbb{E} \{x_i\})^2$. %,  and step (c) holds because of assumptions given in \eqref{eq:assmp2} and  \eqref{eq:assmp4}. 
For the second term in \eqref{eq:lem31},  we have 
%\begin{align}\label{eq:lem34}& 
$\sum_{n=1}^N \alpha_{c_n} \mathbb{E} 
 \bigg\| \sum_{\tau=b_1\kappa_1\kappa_2}^{t-1} \bigg( \sum_{u_m \in \mathcal{C}_n}  \alpha_{u_m}^c\nabla F(\mathbf{w}^{\tau}_{u_m}) - \nabla f_{c_n} (\mathbf{\bar{w}}^{\tau}_{c_n} ) \bigg)\bigg\|^2 \nonumber\\
%\overset{a} \leq & 4 \eta^2 p_s^2 \kappa_1 \kappa_2 \sum_{n=1}^N    \sum_{\tau=b_1\kappa_1\kappa_2}^{t-1} \alpha_{c_n} \mathbb{E} \bigg\| \sum_{u_m \in \mathcal{C}_n }  \alpha_{u_m}^c  \bigg( \nabla F(\mathbf{w}^{\tau}_{u_m}) -  \nabla F (\mathbf{\bar{w}}^{\tau}_{c_n})\bigg) \bigg\|^2 \nonumber\\\overset{b} 
\leq 4 \eta^2 p_s^2  \kappa_1 \kappa_2 L^2  \sum_{m=1}^M  \sum_{\tau=b_1\kappa_1\kappa_2}^{t-1} \alpha_{u_m}
  \mathbb{E} \bigg\|  \mathbf{w}^{\tau}_{u_m}-  \mathbf{\bar{w}}^{\tau}_{c_{u_m}} \bigg\|^2$
%\end{align}
%where step (a) holds 
because $ \| \sum_{i=1}^N x_i  \|^2 \leq N \sum_{i=1}^N \| x_i \|^2$,  and $ \| \sum_{i=1}^N a_i x_i  \|^2 \leq \sum_{i=1}^N a_i \| x_i \|^2 $ with $\sum_{i=1}^N a_i =1, 0 \leq  a_i \leq 1$.
For the third term in \eqref{eq:lem31}, we have
\begin{small}
\begin{align}\label{eq:lem341}
 & 2 \eta^2 p_s^2 \sum_{n=1}^N \alpha_{c_n} \mathbb{E} \bigg\| \sum_{\tau=b_1\kappa_1\kappa_2}^{t-1}  \bigg( \nabla f_{c_{u_m}} (\mathbf{\bar{w}}^{\tau}_{c(u_m)})- \sum_{j=1}^N \alpha_{c_j} \nabla f_{c_j} (\mathbf{\bar{w}}^{\tau}_{c_j})\bigg) \bigg\|^2 \nonumber\\
= & 2 \eta^2  p_s^2 \sum_{n=1}^N  \alpha_{c_n} \mathbb{E} \bigg\| \sum_{\tau=b_1\kappa_1\kappa_2}^{t-1} \bigg( \nabla f_{c_n} (\mathbf{\bar{w}}^{\tau}_{c_n})\pm \nabla f (\mathbf{\bar{w}}^{\tau}_{c_n}) \pm\nabla f (\mathbf{\bar{w}}^{\tau}_{g}) - \sum_{j=1}^N \alpha_{c_j} \nabla f_{c_j} (\mathbf{\bar{w}}^{\tau}_{c_j})\bigg) \bigg\|^2 \nonumber\\
%\leq & 6 \eta^2 p_s^2 \sum_{n=1}^N  \alpha_{c_n} \mathbb{E} \bigg[ \bigg\| \sum_{\tau=b_1\kappa_1\kappa_2}^{t-1} \bigg( \nabla f_{c_n} (\mathbf{\bar{w}}^{\tau}_{c_n}) - \nabla f (\mathbf{\bar{w}}^{\tau}_{c_n})\bigg) \bigg\|^2 + \mathbb{E} \bigg\| \sum_{\tau=b_1\kappa_1\kappa_2}^{t-1} \bigg(  \nabla f (\mathbf{\bar{w}}^{\tau}_{c_n}) - \nabla f (\mathbf{\bar{w}}^{\tau}_{g})\bigg) \bigg\|^2 \nonumber\\
%& +  \mathbb{E} \bigg\| \sum_{\tau=b_1\kappa_1\kappa_2}^{t-1} \bigg(  \nabla f (\mathbf{\bar{w}}^{\tau}_{g}) - \sum_{j=1}^N \alpha_{c_j} \nabla f_{c_j} (\mathbf{\bar{w}}^{\tau}_{c_j})\bigg) \bigg\|^2\bigg]  \nonumber\\
\overset{a} \leq &  12 \eta^2 p_s^2  \kappa_1 \kappa_2 L^2 \sum_{n=1}^N   \sum_{\tau=b_1\kappa_1\kappa_2}^{t-1}\alpha_{c_n} \mathbb{E} \| \mathbf{\bar{w}}^{\tau}_{c_n} -\mathbf{\bar{w}}^{\tau}_{g} \|^2 + 6 p_s^2 \eta^2 \kappa_1^2 \kappa_2^2 \epsilon_g^2 
\end{align}
\end{small}
where step (a) holds because of assumptions of bounded variance of mini-batch gradients and bounded divergences among loss functions and $ \| \sum_{i=1}^N x_i  \|^2 \leq N \sum_{i=1}^N \| x_i \|^2$.

Plugging \eqref{eq:lem33} and \eqref{eq:lem341} back into \eqref{eq:lem31}, and taking average of time, by the help of Lemma 2, if $\eta < \frac{1}{\sqrt{12}L\kappa_1 \kappa_2}$, we can conclude the proof of Lemma \ref{lemma3}.

\subsection{Proof of Theorem \ref{2theorem}}\label{ap:2theorem}
Similar to HFL, the mini-batch gradients, local, cluster and global function in our proposed MACFL algorithm are denoted as $\tilde{g}(\mathbf{w}_{u_m}^t)$, $\tilde{F}(\mathbf{w}_{u_m}^t)$, $\tilde{f}_{c_i}(\mathbf{w}_{c_i}^t)$ and $\tilde{f} (\mathbf{w}_g^t)$, the averaged cluster and global model can be rewritten as $\mathbf{\bar{w}}_{c_n}^{t+1} = \mathbf{\bar{w}}_{c_n}^{t} -\sum_{u_m \in \mathcal{C}_i^t}  \beta_{u_m}^c (\mathbf{\bar{w}}_{c_n}^{t} - \mathbf{w}_{u_m}^{t+1})$, and $\mathbf{\bar{w}}_{g}^{t+1} = \mathbf{\bar{w}}_{g}^{t} - \sum_{i=1}^N \beta_{c_i} (\mathbf{\bar{w}}_{g}^t - \mathbf{\bar{w}}_{c_n}^{t+1} )$. 

In our proposed algorithm, all mobile users can participate in edge model aggregation at each iteration, similar to the proof of Lemma \ref{lemma1},  we have 
\begin{equation} 
 \mathbb{E} \tilde{f}(\mathbf{\bar{w}}^{t+1}_g)
\leq   \mathbb{E} \tilde{f} (\mathbf{\bar{w}}^{t}_g) -\eta\mathbb{E}< \nabla \tilde{f} (\mathbf{\bar{w}}_g^{t}), \sum_{m=1}^M   \beta_{u_m} \tilde{g}(\mathbf{w}_{u_m}^t)> +\frac{\eta^2L}{2} \mathbb{E} \| \sum_{m=1}^M \beta_{u_m} \tilde{g}(\mathbf{w}_{u_m}^t) \|^2
\end{equation}
For the second term, we have 
\begin{small}
\begin{align}
&  -\mathbb{E}< \nabla \tilde{f} (\mathbf{\bar{w}}_g^{t}), \sum_{m=1}^M   \beta_{u_m} \tilde{g}(\mathbf{w}_{u_m}^t)>\nonumber\\
=&  \frac{1}{2} \bigg( \mathbb{E} \| \nabla \tilde{f} (\mathbf{\bar{w}}^{t}_g)-  \sum_{m=1}^M   \beta_{u_m}  \nabla \tilde{F} (\mathbf{w}_{u_m}^t) \|^2  -\mathbb{E} \| \nabla \tilde{f}(\mathbf{\bar{w}}^{t}_g)\|^2- \mathbb{E} \| \sum_{m=1}^M   \beta_{u_m} \nabla \tilde{F}(\mathbf{w}_{u_m}^t) \|^2 \bigg)\nonumber\\
\leq & L^2 \big(\sum_{i=1}^N \beta_{c_i} \mathbb{E} \| \mathbf{\bar{w}}^{t}_g  -\mathbf{\bar{w}}_{c_i}^t \|^2 +  \sum_{m=1}^M \beta_{u_m}  \mathbb{E} \| \mathbf{\bar{w}}_{c_{u_m}}^t - \mathbf{w}_{u_m}^t \|^2\big)
- \frac{1}{2} \big( \mathbb{E} \| \nabla \tilde{f}(\mathbf{\bar{w}}^{t}_g)\|^2%\nonumber\\&
+\mathbb{E} \| \sum_{m=1}^M   \beta_{u_m} \nabla \tilde{F} (\mathbf{w}_{u_m}^t) \|^2\big)
\end{align}
\end{small}
For the third term, we have 
\begin{small}
\begin{align}
\mathbb{E} \| \sum_{m=1}^M \beta_{u_m} \tilde{g}(\mathbf{w}_{u_m}^t) \|^2 
 = &   \bigg(
  \mathbb{E}  \|  \sum_{m=1}^M \beta_{u_m} ( \tilde{g} (\mathbf{w}_{u_m}^t) -\nabla \tilde{F}(\mathbf{w}_{u_m}^t) ) \|^2 
    +    \mathbb{E} \| \sum_{m=1}^M \beta_{u_m} \nabla \tilde{F}(\mathbf{w}_{u_m}^t) \|^2\bigg)   \nonumber\\
\leq &  \sigma_M^2 \sum_{m=1}^M \beta_{u_m}^2  + \mathbb{E} \| \sum_{m=1}^M \beta_{u_m} \nabla \tilde{F}(\mathbf{w}_{u_m}^t) \|^2
\end{align}
\end{small}
where $\theta_M^2$ is the bound of variance of $\tilde{g} (\mathbf{w}_{u_m}^t)$. 

Rearranging the order, dividing both side by $\frac{\eta}{2}$ and taking the average over time, if $\eta \leq \frac{1}{L}$, we have 
\begin{small}
\begin{align}\label{eq:th2ap1}
  \frac{1}{T} \sum_{t=1}^T \mathbb{E} \| \nabla \tilde{f}(\mathbf{\bar{w}}^{t}_g)\|^2   \leq &   \frac{2}{\eta T}  (\mathbb{E} \tilde{f} (\mathbf{\bar{w}}^0_g) - \tilde{f}_{\inf})  + \eta L  \sigma_M^2 \sum_{m=1}^M \beta_{u_m}^2    \nonumber\\ 
 &+  \frac{2 L^2}{T}\sum_{t=0}^{T-1}  \bigg( \sum_{i=1}^N \beta_{c_i} \mathbb{E} \| \mathbf{\bar{w}}^{t}_g - \mathbf{\bar{w}}_{c_i}^t \|^2 +   \sum_{m=1}^M \beta_{u_m}  \mathbb{E} \| \mathbf{\bar{w}}_{c_{u_m}}^t - \mathbf{w}_{u_m}^t \|^2  \bigg)
\end{align}
\end{small}

Similar to Lemma \ref{lemma2}, if $\eta < \frac{1}{\sqrt{12}L\kappa_1}$ , we can obtain the upper bound of the MSE of the local and cluster model parameters as follows:
\begin{small}
 \begin{equation}\label{eq:th2ap2}
 \frac{1}{T} \sum_{t=0}^{T-1} \sum_{m=1}^M  \beta_{u_m} \mathbb{E} \| \mathbf{w}^{t}_{u_m}- \mathbf{\bar{w}}_{c_{u_m}}^t \|^2
 \leq   \frac{ 2 \eta^2 \kappa_1 } {1-12 \eta^2 L^2 \kappa_1^2 }  \bigg( 3 \kappa_1 \epsilon_{M,c}^2 +  \sigma_M^2 \sum_{m=1}^M \beta_{u_m} (1-\beta_{u_m}^c)    \bigg) 
 \end{equation}
\end{small}

Similar to Lemma \ref{lemma3}, if $\eta < \frac{1}{\sqrt{12}L\kappa_1 \kappa_2}$,  we can obtain the upper bound of local model parameters MSE as follows:
\begin{small}
 \begin{align}\label{eq:th2ap3}
  \frac{1}{T} \sum_{t=0}^{T-1}\sum_{i=1}^N \beta_{c_i}  \mathbb{E} \| \mathbf{\bar{w}}^{t}_g - \mathbf{\bar{w}}_{c_i}^t \|^2 
& \leq   \frac{  2 \eta^2 \kappa_1\kappa_2  } {1-12\eta^2L^2 \kappa_1^2 \kappa_2^2 } \bigg[ 3  \kappa_1 \kappa_2  \epsilon_{M,g}^2 +  \frac{ 12 \eta^2  L^2  \kappa_1^3  \kappa_2    } {1-12 \eta^2 L^2 \kappa_1^2  }  \epsilon_{M,c}^2  \nonumber\\ 
&  + 2\sigma_M^2  \sum_{m=1}^M \beta_{u_m} \bigg( (\beta_{u_m}^c-\beta_{u_m}) + \frac{ 2 \eta^2  L^2  \kappa_1^2 \kappa_2  } {1-12 \eta^2 L^2 \kappa_1^2  }  (1-\beta_{u_m}^c) \bigg)  \bigg]
\end{align}
\end{small}
Plugging \eqref{eq:th2ap2} and \eqref{eq:th2ap3} back into \eqref{eq:th2ap1},  if all models are initialized at a same point, and if   $\eta < \frac{1}{\sqrt{12}L\kappa_1 \kappa_2}$, we can conclude the proof of Theorem \ref{2theorem}.

\end{appendix}

%\balance
\bibliographystyle{IEEEtran}
\bibliography{references}

\end{document}